\documentclass[11pt]{article}
\usepackage{booktabs}

\usepackage[preprint]{acl}
\usepackage{multirow}
\usepackage{times}
\usepackage{latexsym}
\usepackage{tabularx}
\usepackage[T1]{fontenc}

\usepackage[utf8]{inputenc}

\usepackage{microtype}

\usepackage{inconsolata}

\usepackage{graphicx}
\usepackage{hyperref}
\usepackage{tcolorbox}
\usepackage{subfigure}

\newtcolorbox{promptbox}{
  colback=gray!5,
  colframe=gray!60,
  boxrule=0.5pt,
  arc=2pt,
  left=6pt,
  right=6pt,
  top=6pt,
  bottom=6pt
}

\usepackage{pifont}

\title{Lost in Execution: On the Multilingual Robustness of Tool Calling \\ in Large Language Models}

\author{
  Zheng Luo\textsuperscript{1} \quad
  T Pranav Kutralingam\textsuperscript{2} \quad
  Ogochukwu N. Okoani\textsuperscript{2} \\
  {\bfseries
    Wanpeng Xu\textsuperscript{2} \quad
    Hua Wei\textsuperscript{2}\footnotemark[1] \quad
    Xiyang Hu\textsuperscript{2}\thanks{Corresponding authors.}
  } \\
  \textsuperscript{1}University of Southern California \quad
  \textsuperscript{2}Arizona State University \\
  \texttt{luozheng@usc.edu}, 
  \texttt{\{tknolast,ookoani,wanpeng.xu,hua.wei,xiyanghu\}@asu.edu}
}

\begin{document}
\maketitle
\begin{abstract}
Large Language Models (LLMs) are increasingly deployed as agents that invoke external tools through structured function calls.
While recent work reports strong tool-calling performance under standard English-centric evaluations, the robustness of tool calling under multilingual user interactions remains underexplored.
In this work, we introduce MLCL, a diagnostic benchmark, and conduct a systematic evaluation of multilingual tool calling across Chinese, Hindi, and the low-resource language Igbo.
Through fine-grained error analysis, we show that many failures occur despite correct intent understanding and tool selection.
We identify \emph{parameter value language mismatch} as a dominant failure mode, where models generate semantically appropriate parameter values in the user's language, violating language-invariant execution conventions.
We further evaluate several inference-time system strategies and find that while these strategies substantially reduce language-induced execution errors, none of them can fully recover English-level performance.

\end{abstract}

%
%


\section{Introduction}


\begin{figure}[t]
    \centering
    \includegraphics[width=\columnwidth]{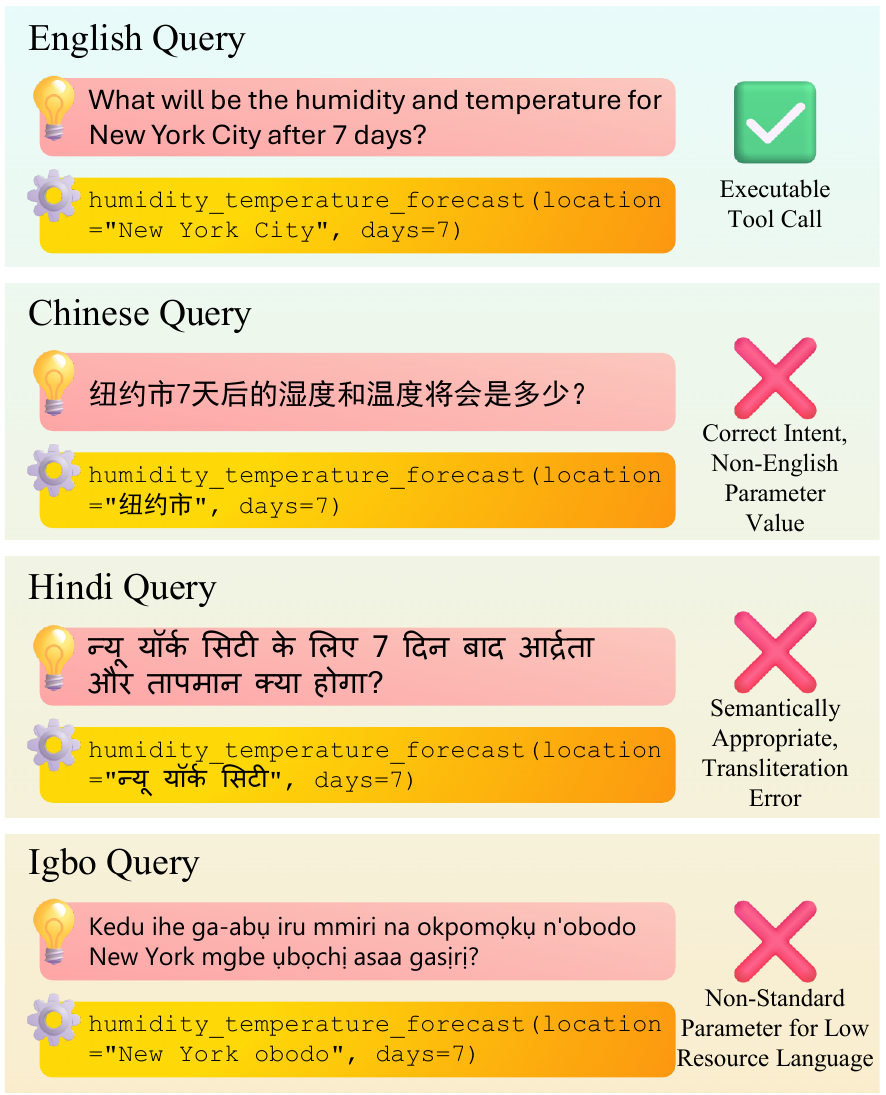}
    \vspace{-6mm}
\caption{ {Multilingual tool-calling failures stem from execution-level parameter mismatches.}
The same user intent expressed in English, Chinese, Hindi, and Igbo yields semantically appropriate tool calls that become non-executable when parameter values violate English-only execution conventions, a failure mode we refer to as \emph{parameter value language mismatch}.}
    \label{fig:intro-example}
\end{figure}

Large Language Models (LLMs) are increasingly deployed as agents that interact with external tools and services, rather than as standalone conversational systems~\cite{parisi2022talmtoolaugmentedlanguage}.
Through tool calling, an LLM can invoke structured APIs to retrieve information, perform computations, or trigger downstream actions, enabling reliable execution beyond free-form text generation~\cite{schick2023toolformer,liu2025uncertainty}.
Accordingly, recent work has focused on improving tool-calling performance through supervised fine-tuning~\cite{tang2023toolalpaca} and reinforcement learning~\cite{qian2025toolrlrewardtoollearning}, and existing benchmarks report strong performance under standard evaluation settings~\cite{chen2024t}.

However, most evaluations implicitly assume that user queries are expressed \emph{only} in English.
In practice, LLM-based agents are often exposed to multilingual user queries while relying on shared, language-invariant tool interfaces.
In such settings, robustness is not only about understanding user intent, but also about maintaining executable behavior at the language--tool boundary.
Despite its practical importance, the impact of linguistic context on tool-calling reliability remains largely unexamined in existing tool-calling benchmarks~\cite{wang2023mint}.

In this work, we show that crossing linguistic boundaries exposes systematic failure patterns that are not captured by standard accuracy metrics.
When user queries are expressed in non-English languages, models frequently generate tool calls that are semantically appropriate yet operationally invalid, as illustrated in Figure~\ref{fig:intro-example}.
In a typical tool-calling setup, an LLM produces a structured function call consisting of a function name (e.g., \texttt{get\_weather()}) and a set of parameter keys (e.g., \textit{``location''} and \textit{``days''}) with values (e.g., \textit{``New York'' and ``7''}) that are passed as arguments to the execution interface.
While the predicted function and the underlying semantics of the arguments are often correct, parameter values are expected to conform to execution-level conventions, such as using English string identifiers.
We refer to failures where models directly copy non-English expressions from the user query into parameter values as \emph{parameter value language mismatch}.
Such mismatches render correct tool calls non-executable, exposing failures that arise at the language--execution boundary rather than from intent misunderstanding.


To systematically study this phenomenon, we introduce a diagnostic multilingual benchmark for tool calling, MLCL, by extending a commonly-used English dataset, the Berkeley Function Calling Leaderboard (BFCL)~\cite{patilberkeley}.
We focus on isolating the effect of query language on tool-calling behavior through controlled query language composition and semantic perturbations, coupled with a fine-grained error taxonomy that distinguishes execution-level violations from semantic errors.
Our evaluation covers Chinese, Hindi, and Igbo, enabling analysis across high-resource and low-resource language settings.

Based on this diagnostic framework, we further examine whether simple inference-time system strategies, including partial translation, explicit prompting, and pre- or post-translation, can mitigate language-induced execution errors.
Across models and languages, these strategies reduce certain error types but fail to consistently recover English-level performance, suggesting that multilingual tool-calling robustness is primarily a system- and interface-level challenge rather than a limitation of intent understanding alone.
Together, our findings highlight the need to better align natural-language interaction with execution conventions in globally deployed LLM-based agents.

In summary, the contributions of this paper are as follows:

\noindent$\bullet$~We introduce MLCL, a diagnostic benchmark for tool-calling robustness under multilingual user queries, covering Chinese, Hindi, and the low-resource language Igbo. The benchmarking dataset characterizes systematic robustness in multilingual tool calling under controlled and interpretable settings. 

\noindent$\bullet$~Through detailed error analysis, we identify \emph{parameter value language mismatch} as a dominant failure mode in multilingual tool calling, despite correct intent understanding and tool selection.

\noindent$\bullet$~We conduct a fine-grained error analysis that separates execution-level violations from semantic errors, revealing systematic differences in error distributions across high-resource and low-resource languages. For high-resource languages like Chinese and Hindi, the major cause of the errors is the tool calling convention and information loss during translation, rather than difficulty in user query comprehension; while for the low-resource Igbo language, the confusion in the user query semantics takes up a larger proportion of errors.

\noindent$\bullet$~We empirically evaluate several simple inference-time system strategies and show that while they substantially reduce language-induced errors, they cannot fully restore English-level performance, highlighting the role of system- and interface-level conventions in multilingual tool calling.


\begin{figure*}[t]
    \centering
    \includegraphics[width=\textwidth]{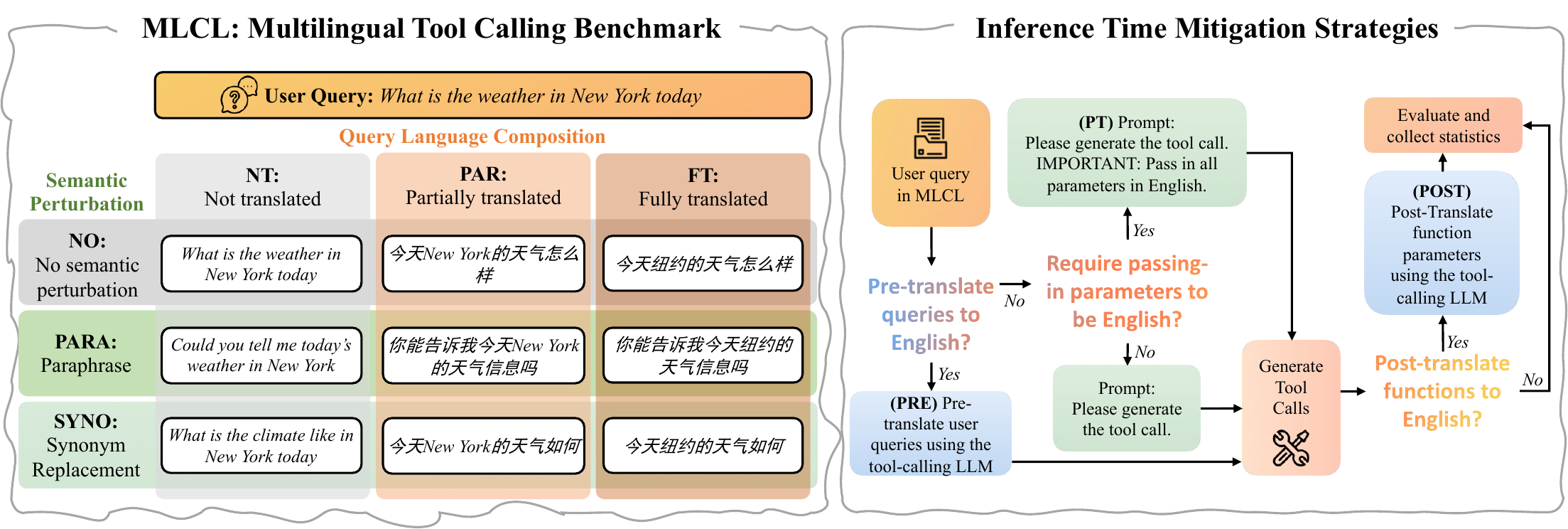}
    \vspace{-8mm}
    \caption{ \textbf{Diagnostic design of the Multilingual Tool-Calling (MLCL) Benchmark and inference-time mitigation strategies.}
The benchmark systematically varies \emph{query language composition} (NT, PAR, FT) and \emph{semantic perturbations} (NO, PARA, SYNO) to isolate multilingual execution failures under a fixed, language-invariant tool interface.
This design exposes cases where tool calls are semantically correct but operationally invalid due to execution-level violations, such as parameter value language mismatch.
The right panel summarizes inference-time mitigation strategies (PT, PRE, POST) evaluated in this work, which reduce some multilingual errors but do not fully eliminate execution-level gaps.}
    \label{fig:tool-benchmark}
    \vspace{-3mm}
\end{figure*}

\section{Related Work}
\paragraph{Tool Calling and Tool Learning}
Recent work has made substantial progress in enabling large language models to interact with external tools through structured APIs.
Early studies, such as Gorilla~\cite{NEURIPS2024_e4c61f57}, highlighted issues such as API hallucination, where models produce syntactically valid but operationally incorrect calls.
To improve reliability, supervised fine-tuning (SFT) approaches~\cite{chen2023fireactlanguageagentfinetuning,zeng-etal-2024-agenttuning,chen-etal-2024-agent,acikgoz2025singlemodelmastermultiturn} and reinforcement learning methods~\cite{qian2025toolrlrewardtoollearning,feng2025retoolreinforcementlearningstrategic,wei2025autotirautonomoustoolsintegrated,wu2025vtoolr1vlmslearnthink,zhang2026selaur,chen2026conformal} have been proposed to align model outputs with tool-calling formats and execution constraints.

Evaluation in this line of work is typically conducted under standardized, predominantly English-language settings~\cite{qin2024toolllm,patilberkeley}.
Benchmarks such as BFCL~\cite{patilberkeley} assess tool selection and parameter accuracy assuming language-consistent execution environments.
While these studies demonstrate strong performance under standard conditions, they largely abstract away linguistic variation in user queries.
As a result, how language differences interact with execution-level conventions in tool calling remains underexplored.
Our work complements this literature by focusing on multilingual robustness and by analyzing execution failures induced by language shifts, rather than proposing new training objectives or architectures.

\paragraph{Robustness in Large Language Models}
Robustness of large language models has been extensively studied under surface-level perturbations such as paraphrasing, synonym substitution, and distributional shifts~\cite{feng2021surveydataaugmentationapproaches,zhou2024surveydataaugmentationlarge,kumar2025robustnesslargelanguagemodels,rabinovich-anaby-tavor-2025-robustness}.
These studies show that even minor linguistic variations can significantly affect model behavior, motivating robustness benchmarks beyond aggregate accuracy metrics.
However, most robustness analyses focus on free-form generation or classification tasks, where failures are primarily semantic.
In contrast, tool calling introduces execution-level constraints: outputs must satisfy strict formatting and parameter conventions to be operationally valid.
Our work extends robustness analysis to this setting by introducing a fine-grained error taxonomy that distinguishes semantic understanding errors from execution-level violations, revealing multilingual failure modes that are obscured by standard accuracy metrics.

\paragraph{Multilingual Evaluation of Language Models}
Multilingual evaluation benchmarks assess cross-lingual understanding, reasoning, and instruction following across diverse languages~\cite{ruder-etal-2021-xtreme,ahuja-etal-2023-mega,shi2023language,chen-etal-2024-breaking,xu2026languageshapesmentalhealth,zhou2026fairnessfluencyinvestigationlanguage}.
While these studies show that large language models can preserve semantic reasoning across languages, prior work on multilingual settings with structured outputs suggests that maintaining cross-lingual consistency remains challenging even when translation quality is high~\cite{moradshahi-etal-2023-x}.
However, existing multilingual evaluations rarely consider scenarios where model outputs must interface with external, language-sensitive systems such as tools or APIs.
Our work addresses this gap by studying multilingual robustness in tool calling and identifying \emph{parameter value language mismatch} as a distinct execution-level failure mode.

\section{Multilingual Tool-Calling Benchmark}
\label{sec:benchmark}

To characterize multilingual tool-calling failures under controlled, interpretable settings,
rather than treating multilinguality as a simple dataset extension, we introduce a diagnostic benchmark focusing on failure modes that arise at the interface between natural-language input and a language-invariant execution environment (Figure~\ref{fig:tool-benchmark}).

\subsection{Design Goals and Diagnostic Scope}
Our design goal is to separate errors caused by \emph{query-language variation} from those caused by \emph{execution-interface conventions}.
To this end, the benchmark dataset is structured along two orthogonal diagnostic dimensions:
\emph{Query Language Composition} and \emph{Semantic Perturbation Design}.
Query Language Composition controls how non-English content is introduced in the user query, while Semantic Perturbation Design applies surface-form variations that preserve intent.
Together, these dimensions define a compact diagnostic space for attributing multilingual tool-calling failures to language understanding, execution-interface mismatches, or parameter realization.

\subsection{Dataset Construction}
Our benchmark is constructed by extending an existing English tool-calling dataset into a multilingual diagnostic suite, while keeping the execution interface fixed.
This design ensures that any observed performance degradation can be attributed to changes in the natural-language input, rather than differences in tool schemas, APIs, or evaluation criteria.

\subsubsection{Base Tasks and Execution Interface}
We formulate tool calling as a structured generation problem at the interface between natural-language understanding and programmatic execution.
Each task consists of a user query and a set of candidate tools, where each tool is specified by a function name and a fixed parameter schema.

Given a query and tool descriptions, the model generates a tool call by selecting a function name and producing concrete parameter values.
Correct execution requires adherence to execution-level constraints such as exact parameter keys (e.g., \textit{``location''} in Figure~
\ref{fig:intro-example}) and surface-form conventions for parameter values (e.g., \textit{``New York''} in Figure~
\ref{fig:intro-example}).
The execution interface is kept language-invariant and English-only throughout this work.
The original English queries define the \textbf{NT} (Not Translated) reference setting.
All the later multilingual variants modify only the natural-language query while keeping the execution interface fixed.

\subsubsection{Query Language Composition}
To model multilingual user interactions, we construct translated variants of each query while preserving the execution interface.
Only the natural-language query text is modified; function names, parameter keys, and tool descriptions are not translated.
This isolates the effect of query language on tool selection and parameter realization.

In the \textbf{FT} (Fully Translated) setting, the entire user query is translated into a target language.
In the \textbf{PAR} (Partially Translated) setting, ground-truth parameter values remain in English while the surrounding context is translated, yielding mixed-language queries that reduce parameter-language mismatch while preserving multilingual context.
Translations are generated by GPT-5 and manually verified to ensure semantic equivalence with the original English queries, as detailed in 
Appendix~\ref{sec:verification}.
Verification focuses on intent preservation rather than literal word-level correspondence, reflecting realistic multilingual usage.

\subsubsection{Semantic Perturbation Design}
To assess whether multilingual execution failures are sensitive to benign surface-form variation, we introduce semantic perturbations that preserve intent while modifying wording.
These perturbations are applied consistently across the English and translated datasets.
We consider two perturbation types:
\textbf{PARA} generates paraphrased versions of each query with altered phrasing but unchanged meaning.
\textbf{SYNO} applies synonym substitutions to individual words where appropriate.
Both perturbations are generated using GPT-5 and manually reviewed to maintain semantic consistency.
For partially translated queries, perturbations are applied without explicitly protecting English parameter strings, allowing semantic variation to naturally interact with mixed-language inputs.

\subsection{Experimental Protocol}
This section describes how the benchmark defined above is instantiated for evaluation. It varies only the query language composition and semantic perturbations, enabling controlled comparison across languages, models, and settings.

\paragraph{Base Dataset}
We adopt the BFCL V4~\cite{patilberkeley} as the base benchmark in English since it provides tool-calling tasks with strict execution-level ground truth, making it well-suited for analyzing failures that arise from parameter realization and interface compliance.
To avoid confounding factors such as dialogue context or live execution, we use the \texttt{BFCL\_v4\_multiple.json} subset, which contains single-turn queries paired with multiple candidate functions.

\paragraph{Language Evaluated}
Besides English, we evaluate three languages with distinct linguistic properties and resource availability: Chinese, Hindi, and Igbo.
Chinese represents a high-resource language with logographic writing, Hindi introduces richer morphology and more flexible word order, and Igbo serves as a low-resource language with limited representation in tool-calling training data.
This selection is intended to test whether observed multilingual failure patterns generalize across typologically diverse languages.

\paragraph{Models Evaluated}
We evaluate a diverse set of large language models with explicit tool-calling capabilities, including both proprietary and open-source systems.
The selected models span multiple model families and scales, allowing us to examine whether multilingual tool-calling robustness varies with architecture and capacity.
All models are evaluated using their officially supported tool-calling interfaces and recommended decoding configurations.
Model-specific input and output formats are summarized in Appendix~\ref{sec:official}.

\begin{table}[t!]
\centering
\small
\caption{ {Models evaluated in the multilingual tool-calling benchmark.}
We include proprietary and open-source models from multiple families and scales to ensure that observed multilingual tool-calling failures are evaluated across diverse model architectures.}
\label{tab:models-table}
  \vspace{-3mm}
\begin{tabular}{ll}
\toprule
Model Family & Model Name \\ \toprule
GPT-5 & GPT-5, GPT-5 mini, GPT-5 nano \\ \hline
DeepSeek & DeepSeek V3.2 \\ \hline
\multirow{2}{*}{Llama 3.1} & meta-llama/Llama-3.1-8B-Instruct \\ \cline{2-2} 
 & meta-llama/Llama-3.1-70B-Instruct \\ \hline
\multirow{5}{*}{Qwen 3} & Qwen/Qwen3-8B \\ \cline{2-2} 
 & Qwen/Qwen3-14B \\ \cline{2-2} 
 & Qwen/Qwen3-30B-A3B \\ \cline{2-2} 
 & Qwen/Qwen3-32B \\ \cline{2-2} 
 & Qwen/Qwen3-Next-80B-A3B-Instruct \\ \hline
\multirow{2}{*}{Granite 4} & ibm-granite/granite-4.0-h-tiny \\ \cline{2-2} 
 & ibm-granite/granite-4.0-h-small \\ \bottomrule
\end{tabular}
\end{table}

\paragraph{Evaluation Metrics and Error Attribution}
Evaluation follows the BFCL protocol~\cite{patilberkeley}, which requires exact matching of function names, parameter keys, and parameter values.
While strict surface-form matching penalizes benign variation, it directly reflects whether a generated tool call can be executed without additional system intervention.

\begin{figure}[t!]
    \centering
    \includegraphics[width=\linewidth,
  trim=1mm 0mm 1mm 0mm,
  clip]{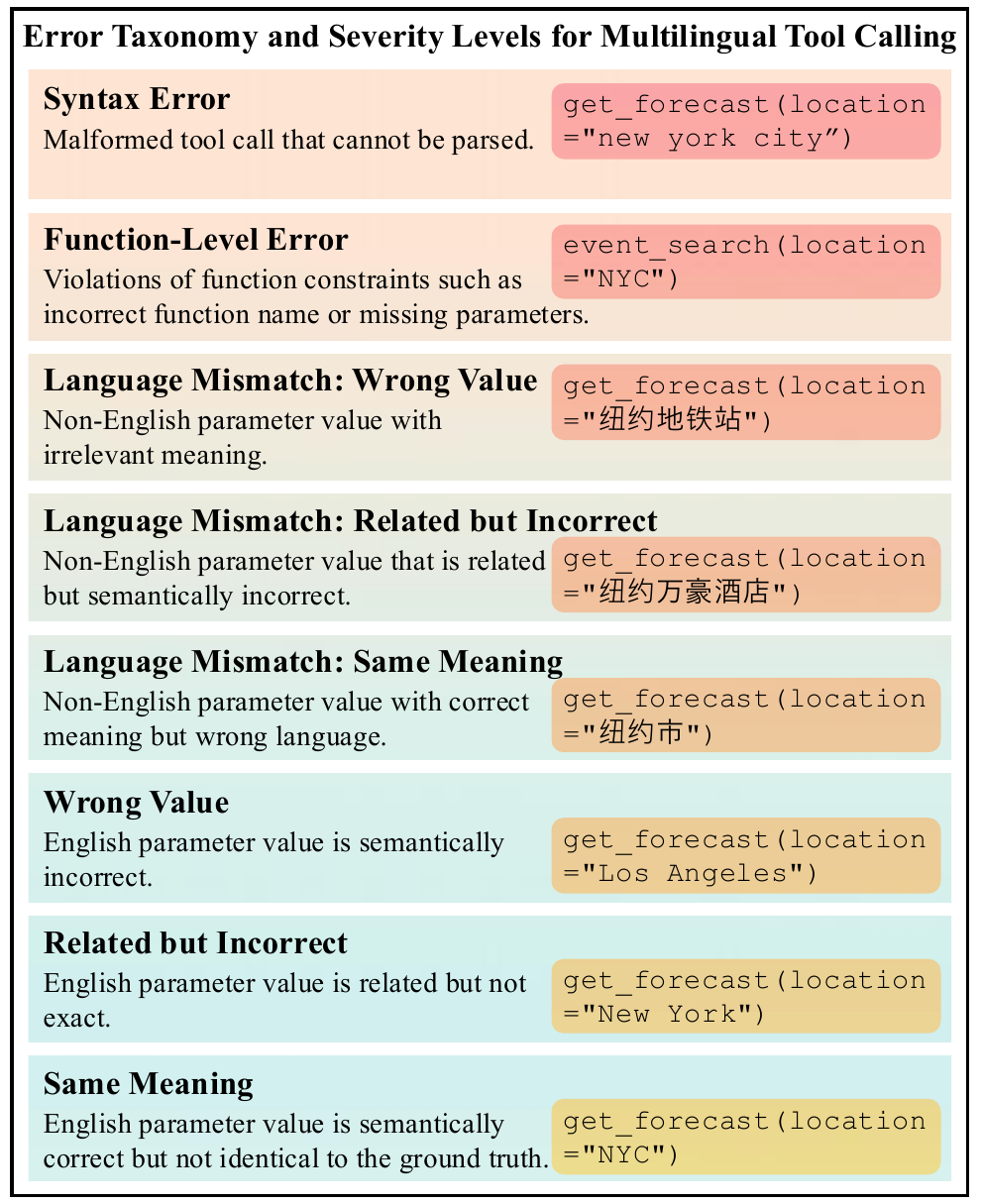}
      \vspace{-8mm}
    \caption{ {Error taxonomy and severity levels used for evaluating multilingual tool calling}.
Error categories are ordered from most severe (top) to least severe (bottom),
with illustrative examples for each category.}
    \label{fig:error-tax}
\end{figure}

To support multilingual analysis, we extend BFCL’s evaluation by explicitly separating semantic correctness from language conformity in parameter values. As illustrated in Figure~\ref{fig:error-tax}, errors are organized by severity.
At the most severe level are syntax and function-level errors, which prevent execution due to malformed outputs or schema violations.
A central focus of this work is \emph{parameter value language mismatch}, where parameter values are generated in a non-English language despite correct intent understanding and tool selection.
We further distinguish these cases by semantic correctness, separating execution failures caused purely by language mismatch from those involving incorrect values.
This allows us to distinguish errors caused by incorrect intent understanding or argument selection from those arising solely due to violations of execution-level language conventions.
Results are reported using \textbf{overall error rate} together with a structured \textbf{error breakdown}, enabling fine-grained analysis across query language composition and semantic perturbation settings.
Detailed definitions and examples for each category are provided in Appendix~\ref{app:sec:error}.


\begin{figure*}[t!]
  \centering
  \includegraphics[width=\textwidth]{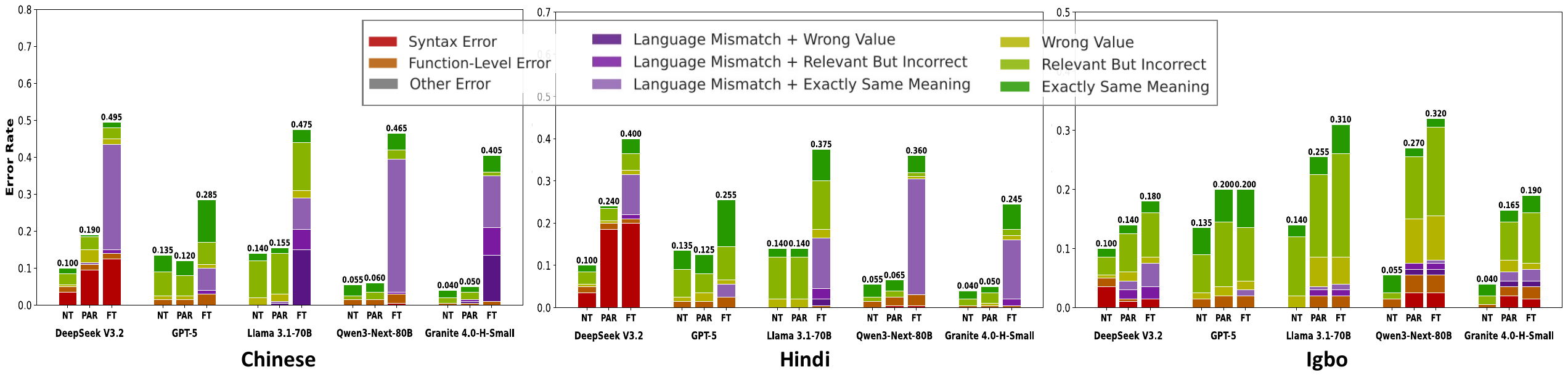}
  \vspace{-8mm}
\caption{ {Error distributions for five representative models under English (NT), partially translated (PAR), and fully translated (FT) queries in Chinese, Hindi, and Igbo.}
Across languages and model families, moving from NT to FT systematically increases execution-level errors, driven primarily by \emph{parameter value language mismatch} (in purple color), while PAR substantially reduces these violations.
The consistency of this trend across models indicates a shared failure mechanism at the language--execution interface rather than model-specific weaknesses.}

  \label{fig:panta-multilang}
\end{figure*}

\subsection{Results}
\label{sec:exp:benchmarking-results}
We analyze multilingual tool-calling behavior across query language composition and semantic perturbation settings, using the execution-oriented error taxonomy illustrated in Figure~\ref{fig:error-tax}.
Results are reported in terms of error composition and severity rather than overall accuracy, reflecting whether failures arise from execution-interface violations or semantic misunderstanding.

\vspace{1mm}
\noindent\textbf{Introducing fully non-English queries substantially increases execution-level errors.}
Figure~\ref{fig:panta-multilang} shows the error breakdown for five representative models when moving from the English reference setting (NT) to fully translated queries (FT).
Across all three languages, the FT setting leads to a pronounced increase in execution failures, dominated by parameter value language mismatch.
In the FT setting, models frequently copy non-English tokens from the user query directly into parameter values, violating the English-only execution interface.
In most cases, these values remain semantically correct, indicating that intent understanding and tool selection have succeeded and that failures arise primarily at the language--execution boundary.

\vspace{1mm}
\noindent\textbf{The composition of multilingual tool-calling errors differs systematically across languages.}
Although the increase in execution-level errors under FT is consistent across languages, the dominant error types vary.
As shown in Figure~\ref{fig:panta-multilang}, parameter value language mismatch is most prevalent for Chinese, followed by Hindi, and is least frequent for Igbo.
This pattern suggests that models are more likely to reuse query tokens from high-resource languages as parameter values, while avoiding doing so for lower-resource languages.
As a result, lower-resource languages exhibit fewer language-mismatch errors but a higher proportion of errors related to semantic misunderstanding, indicating that execution-level mismatch and language comprehension contribute differently across languages.

\vspace{1mm}
\noindent\textbf{Partial translation isolates execution-interface violations from language understanding errors.}
For several models, including GPT-5 and Llama~3.1--70B, the partially translated (PAR) setting exhibits fewer execution-level errors than FT setting, and in some cases matches or outperforms the English reference.
By preserving English parameter strings while translating the surrounding context, PAR removes parameter value language mismatch without substantially altering the semantic content of the query.
These results indicate that, once execution-interface violations are controlled for, many models can interpret non-English queries with comparable reliability to English ones.

\begin{figure*}[t]
  \centering
  \includegraphics[width=\textwidth]{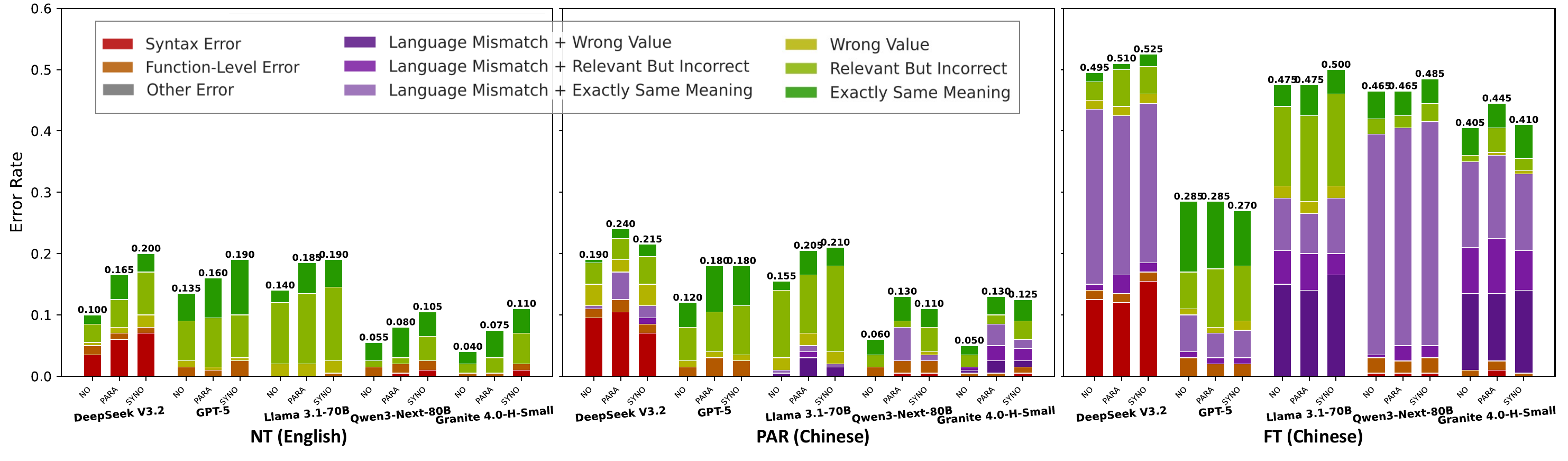}
  \vspace{-8mm}
\caption{ {Evaluation of semantic perturbations for five representative models on the Chinese dataset} under paraphrasing (PARA) and synonym substitution (SYNO) across NT, PAR, and FT settings.
Semantic perturbations substantially increase errors when exact English parameter surface forms are required (NT), but have limited additional impact in fully translated (FT) queries dominated by parameter value language mismatch (in purple color).
The partially translated (PAR) setting exhibits intermediate sensitivity, indicating an interaction between semantic variation and execution-level language constraints.}
  \label{fig:by_noise_chinese}
\vspace{-3mm}
\end{figure*}

\vspace{1mm}
\noindent\textbf{Semantic perturbations exacerbate execution errors when strict surface-form matching is required.}
Figure~\ref{fig:by_noise_chinese} summarizes the impact of paraphrasing (PARA) and synonym substitution (SYNO) across query language composition settings. Additional results can be found in Appendix~\ref{app:sec:semantic}.
In the English reference setting, semantic perturbations substantially increase execution failures because altered surface forms no longer match expected parameter values.
In contrast, semantic perturbations have limited additional effect in fully translated settings, where execution errors are already dominated by language mismatch.
Partially translated settings exhibit intermediate sensitivity: perturbations can replace preserved English parameter strings with non-English equivalents, reintroducing execution-level violations.
Overall, semantic noise amplifies multilingual tool-calling failures primarily when strict surface-form conformity is required.

\section{Inference Time Mitigation Strategies}

\subsection{Motivation and Scope}
As shown in Section~\ref{sec:exp:benchmarking-results}, most multilingual tool-calling failures originate from execution-level language mismatch rather than semantic misunderstanding: Models often select the correct tool and generate semantically appropriate arguments, but fail to conform to the English-only execution interface when queries are non-English.
This raises a practical question: \emph{can simple inference-time strategies reduce such failures without retraining or fine-tuning models?}
We focus on lightweight interventions that are compatible with deployed systems and do not require changes to model weights.
All mitigation strategies are evaluated under the same task abstraction and error taxonomy as in the main benchmark, enabling direct comparison with the NT, PAR, and FT settings.

\subsection{Mitigation Methods}
We consider three lightweight inference-time mitigation strategies that target execution-level language mismatch at different stages of the tool-calling pipeline, as shown in Figure~\ref{fig:tool-benchmark}.
All strategies operate without modifying model parameters and are compatible with deployed systems. 
Implementation details, prompts, and translation procedures for all mitigation strategies are provided in Appendix~\ref{sec:mitigation-details}.

\noindent$\bullet$~\textbf{Prompt-Level Instruction (PT)}
PT fully translates the user query and explicitly instructs the model to output parameter values in English.
This strategy tests whether execution-interface conventions can be enforced through natural-language instructions alone.

\noindent$\bullet$~\textbf{Pre-Translation of User Queries (PRE)}
PRE translates non-English user queries into English before tool calling.
By normalizing the input, this strategy removes multilingual variation from the tool-calling step and serves as an upper bound on mitigation through input preprocessing.

\begin{figure}[t]
  \centering
  \includegraphics[width=\columnwidth]{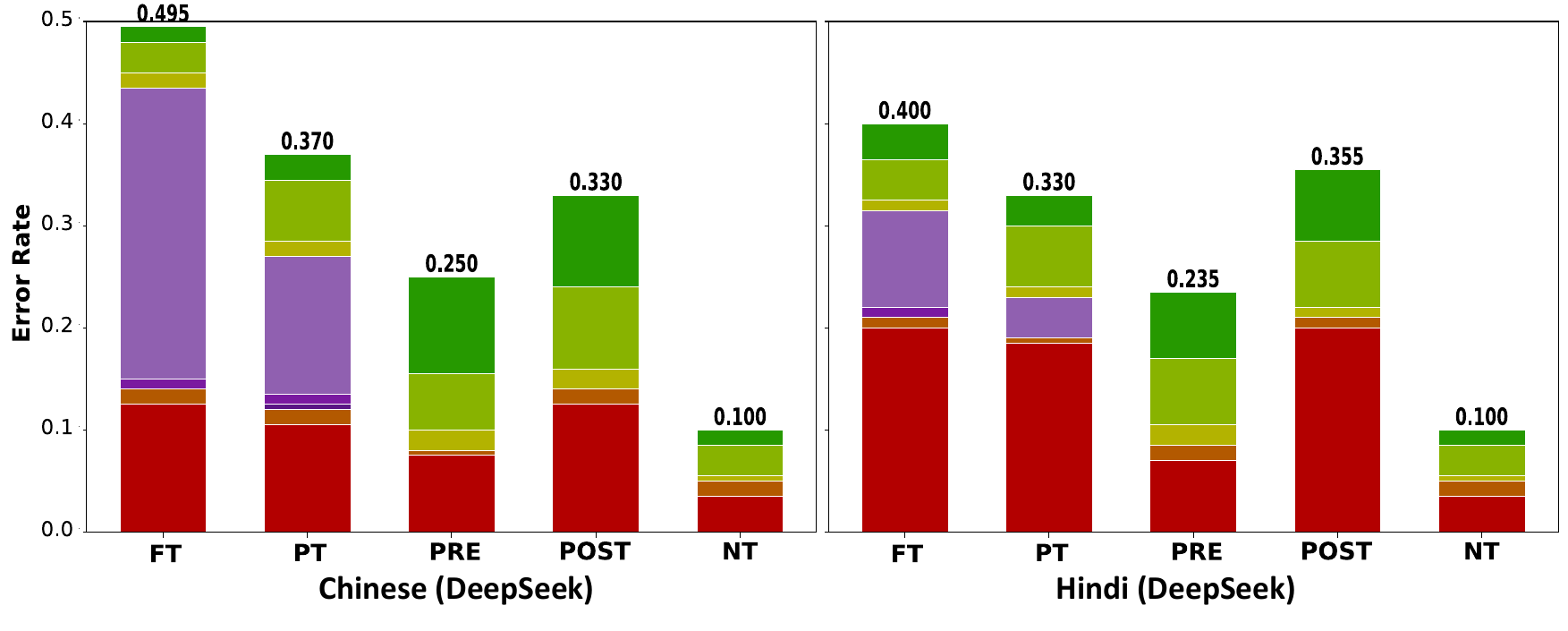}
  \vspace{-8mm}
\caption{ {Evaluation of inference-time mitigation strategies for DeepSeek~V3.2 on Chinese and Hindi datasets}, including explicit prompting (PT), pre-translation (PRE), and post-translation (POST).
Across both languages, these strategies reduce parameter value language mismatch errors but fail to recover English-only (NT) performance.
PRE generally outperforms POST due to access to the full query context, while PT exhibits inconsistent compliance, indicating persistent execution-level gaps beyond simple translation fixes.}
  \label{fig:deepseek-chinese-hindi-mitigation}
\vspace{-3mm}
\end{figure}

\noindent$\bullet$~\textbf{Post-Translation of Parameter Values (POST)}
POST translates generated parameter values into English after tool calling but before execution.
This strategy directly targets parameter value language mismatch while preserving the model’s original tool selection and argument structure.

\subsection{Evaluation of Mitigation Strategies}
\paragraph{Inference-time mitigation strategies reduce execution-interface violations but do not recover English-level performance.}
Figure~\ref{fig:deepseek-chinese-hindi-mitigation} shows representative results on Chinese queries with different mitigation strategies. Similar trends are also observed across different models and languages. 
Prompt-level instruction (PT) reduces parameter value language mismatch relative to the fully translated setting, but compliance is inconsistent, and residual mismatches remain common.
Pre-translation (PRE) and post-translation (POST) further reduce language mismatch errors, with PRE generally more effective due to its access to the full query context.
However, none of these strategies eliminates execution failures.
Translation introduces semantic drift and surface-form normalization, replacing language mismatch with new execution-level errors under strict matching.
We include a case study in Appendix~\ref{app:sec:case} for better understanding.
%



\begin{figure}[t]
  \centering
  \includegraphics[width=\columnwidth]{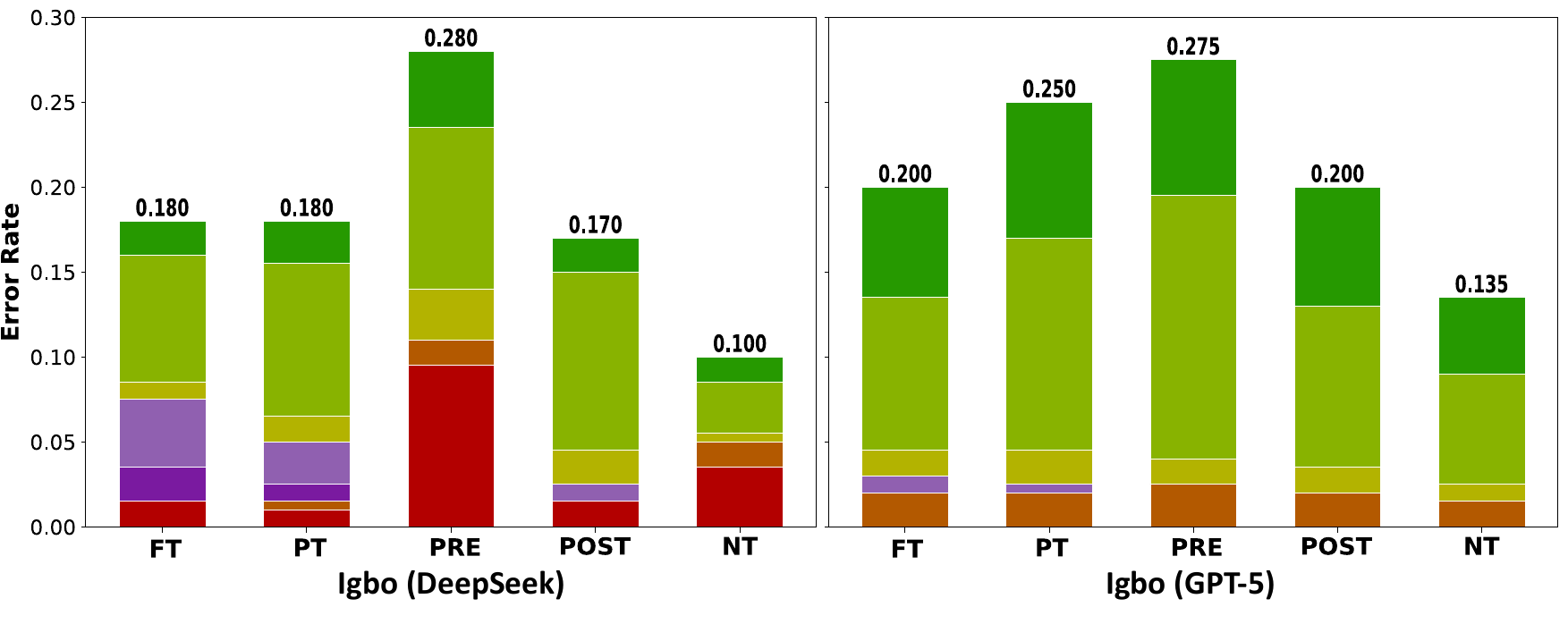}
  \vspace{-8mm}
\caption{ {Inference-time mitigation effects on a low-resource language (Igbo) for DeepSeek~V3.2 and GPT-5}.
Unlike Chinese and Hindi, translation-based mitigations (PT, PRE, POST) provide limited benefit and can increase errors, as parameter value language mismatch is rare even without explicit prompting.
Instead, remaining failures are dominated by query understanding errors, indicating that mitigation strategies targeting execution-level language mismatches are less effective for low-resource languages.}
  \label{fig:deepseek-igbo-mitigation}
\vspace{-3mm}
\end{figure}

\paragraph{Mitigation behavior differs systematically for low-resource languages.}
Low-resource languages such as Igbo exhibit mitigation behavior that differs from that of higher-resource languages like Chinese and Hindi.
As shown in Figure~\ref{fig:deepseek-igbo-mitigation}, translation-based mitigation strategies often provide limited benefit for Igbo and can even degrade performance.
Unlike higher-resource languages, parameter value language mismatch is rare for Igbo even without explicit prompting.
This suggests that models are less likely to directly copy low-resource language tokens from the query into parameter values, thereby reducing execution-interface violations.
As a result, mitigation strategies that primarily target language mismatch, such as PT and POST, have limited room to improve performance.
The remaining failures are instead dominated by query understanding errors, including unstable semantic grounding and imprecise mapping between natural-language expressions and the target parameter schema.


\section{Discussion}
This work analyzes multilingual tool-calling failures under controlled execution settings and demonstrates that the multilingual degradation is primarily driven by execution-interface violations rather than semantic misunderstandings.

\textbf{Execution interfaces are a central bottleneck.}
Across models and languages, non-English queries increase execution-level errors, most notably parameter value language mismatch.
In many cases, models select the correct tool and generate semantically appropriate arguments, yet fail to satisfy strict surface-form requirements imposed by the execution interface.
This indicates a systematic mismatch between flexible natural-language generation and rigid programmatic constraints.

\textbf{Inference-time mitigation is helpful but insufficient.}
Prompting and translation-based strategies can reduce language mismatch errors, but none recover English-level performance.
While these methods alleviate direct language violations, they introduce semantic drift and normalization effects that lead to new execution failures.
This suggests that inference-time patching can mitigate symptoms but cannot fully resolve the underlying issue.

\textbf{Implications for deployment with global users.}
For real-world systems, these results imply that multilingual users may experience reliability gaps even when intent understanding succeeds.
Without explicit handling of execution-interface constraints, tool-calling LLMs risk failures for non-English queries, limiting global deployment robustness.

\vspace{2mm}
\noindent Overall, our findings suggest that improving multilingual tool calling requires not only stronger language understanding, but also execution-aware system design that aligns natural-language variability with programmatic interfaces.

\section{Conclusion}
This paper investigates multilingual tool calling under controlled execution settings and shows that performance degradation is driven primarily by execution-interface violations rather than failures in intent understanding.
Across models and languages, fully translated queries substantially increase execution errors, with parameter value language mismatch emerging as a dominant failure mode.
We further show that inference-time mitigation strategies can reduce specific classes of errors but do not recover English-level performance, often introducing new execution failures through semantic drift or surface-form variation.
These findings highlight multilingual tool calling as a system-level challenge that cannot be addressed by inference-time interventions alone and motivate more careful alignment between language generation and execution interfaces.

\section*{Acknowledgment}
The work was partially supported by NSF award \#2442477 and \#2550203. We thank Amazon Research Awards, Cisco Faculty Research Awards, and Toyota Faculty Research Awards. The authors acknowledge Google and OpenAI for providing us with API credits and Research Computing at Arizona State University for providing computing resources. 
The views and conclusions in this paper are those of the authors and should not be interpreted as representing any funding agencies.

\section*{Limitations}

This study has several limitations that should be considered when interpreting the results.
First, our evaluation is based on the vanilla subset of the Berkeley Function Call Leaderboard to focus on single-turn tool-calling scenarios with predefined function interfaces.
This choice allows us to isolate execution-level effects under controlled conditions, but the results should be interpreted within this single-turn setting.
Second, although we examine three typologically diverse languages, including a low-resource language, the set of languages remains limited and does not cover all linguistic families or writing systems.
Our goal is not exhaustive multilingual coverage, but to identify systematic patterns that emerge across representative high-resource and low-resource languages.
Third, our definition of execution correctness assumes that tool interfaces expect parameter values in English.
This reflects a common design choice in current tool-calling systems and benchmarks, and our analysis focuses on understanding model behavior under this convention rather than advocating a particular interface design.
Fourth, the inference-time strategies studied in this paper are intentionally simple.
They are used as diagnostic probes to help identify the sources of multilingual failures, rather than as optimized mitigation methods.
Finally, although we analyze model scale and architecture effects across several model families, our conclusions are constrained by the specific models evaluated and should not be interpreted as claims about all future model designs.

\section*{Ethical Considerations}

This work evaluates multilingual tool calling by LLMs under a controlled, English-only execution interface, using translated and perturbed versions of existing benchmark queries (Chinese, Hindi, Igbo) without collecting user data or conducting human-subject studies. The primary risk is that tool-calling failures in non-English settings can cause silent non-executable calls or incorrect downstream actions, which may disproportionately affect non-English users; our contribution is a diagnostic taxonomy and mitigation analysis intended to help system builders detect and reduce such execution-level errors, although stronger tool-calling can also increase the effectiveness of automation if connected to high-impact tools. Limitations include partial language coverage and possible translation artifacts that can introduce semantic drift; we report language-specific error patterns to avoid overgeneralization. AI-assisted tools were used to improve the writing of this paper.

\bibliography{custom}

@misc{xu2026languageshapesmentalhealth,
      title={Language Shapes Mental Health Evaluations in Large Language Models}, 
      author={Jiayi Xu and Xiyang Hu},
      year={2026},
      eprint={2603.06910},
      archivePrefix={arXiv},
      primaryClass={cs.CL},
      url={https://arxiv.org/abs/2603.06910}, 
}

@article{zhang2026selaur,
  title={SELAUR: Self Evolving LLM Agent via Uncertainty-aware Rewards},
  author={Zhang, Dengjia and Liu, Xiaoou and Cheng, Lu and Wang, Yaqing and Murray, Kenton and Wei, Hua},
  journal={arXiv preprint arXiv:2602.21158},
  year={2026}
}

@inproceedings{liu2025uncertainty,
  title={Uncertainty quantification and confidence calibration in large language models: A survey},
  author={Liu, Xiaoou and Chen, Tiejin and Da, Longchao and Chen, Chacha and Lin, Zhen and Wei, Hua},
  booktitle={Proceedings of the 31st ACM SIGKDD Conference on Knowledge Discovery and Data Mining V. 2},
  pages={6107--6117},
  year={2025}
}

@article{chen2026conformal,
  title={Conformal Feedback Alignment: Quantifying Answer-Level Reliability for Robust LLM Alignment},
  author={Chen, Tiejin and Liu, Xiaoou and Nandam, Vishnu and Liou, Kuan-Ru and Wei, Hua},
  journal={arXiv preprint arXiv:2601.17329},
  year={2026}
}

@misc{zhou2026fairnessfluencyinvestigationlanguage,
      title={Fairness or Fluency? An Investigation into Language Bias of Pairwise LLM-as-a-Judge}, 
      author={Xiaolin Zhou and Zheng Luo and Yicheng Gao and Qixuan Chen and Xiyang Hu and Yue Zhao and Ruishan Liu},
      year={2026},
      eprint={2601.13649},
      archivePrefix={arXiv},
      primaryClass={cs.CL},
      url={https://arxiv.org/abs/2601.13649}, 
}

@inproceedings{chen-etal-2024-breaking,
    title = "Breaking Language Barriers in Multilingual Mathematical Reasoning: Insights and Observations",
    author = "Chen, Nuo  and
      Zheng, Zinan  and
      Wu, Ning  and
      Gong, Ming  and
      Zhang, Dongmei  and
      Li, Jia",
    editor = "Al-Onaizan, Yaser  and
      Bansal, Mohit  and
      Chen, Yun-Nung",
    booktitle = "Findings of the Association for Computational Linguistics: EMNLP 2024",
    month = nov,
    year = "2024",
    address = "Miami, Florida, USA",
    publisher = "Association for Computational Linguistics",
    url = "https://aclanthology.org/2024.findings-emnlp.411/",
    doi = "10.18653/v1/2024.findings-emnlp.411",
    pages = "7001--7016"
}

@inproceedings{ahuja-etal-2023-mega,
    title = "{MEGA}: Multilingual Evaluation of Generative {AI}",
    author = "Ahuja, Kabir  and
      Diddee, Harshita  and
      Hada, Rishav  and
      Ochieng, Millicent  and
      Ramesh, Krithika  and
      Jain, Prachi  and
      Nambi, Akshay  and
      Ganu, Tanuja  and
      Segal, Sameer  and
      Ahmed, Mohamed  and
      Bali, Kalika  and
      Sitaram, Sunayana",
    editor = "Bouamor, Houda  and
      Pino, Juan  and
      Bali, Kalika",
    booktitle = "Proceedings of the 2023 Conference on Empirical Methods in Natural Language Processing",
    month = dec,
    year = "2023",
    address = "Singapore",
    publisher = "Association for Computational Linguistics",
    url = "https://aclanthology.org/2023.emnlp-main.258/",
    doi = "10.18653/v1/2023.emnlp-main.258",
    pages = "4232--4267"
}

@inproceedings{ruder-etal-2021-xtreme,
    title = "{XTREME}-{R}: Towards More Challenging and Nuanced Multilingual Evaluation",
    author = "Ruder, Sebastian  and
      Constant, Noah  and
      Botha, Jan  and
      Siddhant, Aditya  and
      Firat, Orhan  and
      Fu, Jinlan  and
      Liu, Pengfei  and
      Hu, Junjie  and
      Garrette, Dan  and
      Neubig, Graham  and
      Johnson, Melvin",
    editor = "Moens, Marie-Francine  and
      Huang, Xuanjing  and
      Specia, Lucia  and
      Yih, Scott Wen-tau",
    booktitle = "Proceedings of the 2021 Conference on Empirical Methods in Natural Language Processing",
    month = nov,
    year = "2021",
    address = "Online and Punta Cana, Dominican Republic",
    publisher = "Association for Computational Linguistics",
    url = "https://aclanthology.org/2021.emnlp-main.802/",
    doi = "10.18653/v1/2021.emnlp-main.802",
    pages = "10215--10245"
}

@misc{qian2025toolrlrewardtoollearning,
      title={ToolRL: Reward is All Tool Learning Needs}, 
      author={Cheng Qian and Emre Can Acikgoz and Qi He and Hongru Wang and Xiusi Chen and Dilek Hakkani-Tür and Gokhan Tur and Heng Ji},
      year={2025},
      eprint={2504.13958},
      archivePrefix={arXiv},
      primaryClass={cs.LG},
      url={https://arxiv.org/abs/2504.13958}, 
}

@misc{feng2025retoolreinforcementlearningstrategic,
      title={ReTool: Reinforcement Learning for Strategic Tool Use in LLMs}, 
      author={Jiazhan Feng and Shijue Huang and Xingwei Qu and Ge Zhang and Yujia Qin and Baoquan Zhong and Chengquan Jiang and Jinxin Chi and Wanjun Zhong},
      year={2025},
      eprint={2504.11536},
      archivePrefix={arXiv},
      primaryClass={cs.CL},
      url={https://arxiv.org/abs/2504.11536}, 
}

@inproceedings{rabinovich-anaby-tavor-2025-robustness,
    title = "On the Robustness of Agentic Function Calling",
    author = "Rabinovich, Ella  and
      Anaby Tavor, Ateret",
    editor = "Cao, Trista  and
      Das, Anubrata  and
      Kumarage, Tharindu  and
      Wan, Yixin  and
      Krishna, Satyapriya  and
      Mehrabi, Ninareh  and
      Dhamala, Jwala  and
      Ramakrishna, Anil  and
      Galystan, Aram  and
      Kumar, Anoop  and
      Gupta, Rahul  and
      Chang, Kai-Wei",
    booktitle = "Proceedings of the 5th Workshop on Trustworthy NLP (TrustNLP 2025)",
    month = may,
    year = "2025",
    address = "Albuquerque, New Mexico",
    publisher = "Association for Computational Linguistics",
    url = "https://aclanthology.org/2025.trustnlp-main.20/",
    doi = "10.18653/v1/2025.trustnlp-main.20",
    pages = "298--304",
    ISBN = "979-8-89176-233-6"
}

@inproceedings{patilberkeley,
  title={The Berkeley Function Calling Leaderboard (BFCL): From Tool Use to Agentic Evaluation of Large Language Models}, 
  author={Patil, Shishir G. and Mao, Huanzhi and Cheng-Jie Ji, Charlie and Yan, Fanjia and Suresh, Vishnu and Stoica, Ion and E. Gonzalez, Joseph},
  booktitle={Forty-second International Conference on Machine Learning},
  year={2025},
}

@misc{kumar2025robustnesslargelanguagemodels,
      title={Robustness in Large Language Models: A Survey of Mitigation Strategies and Evaluation Metrics}, 
      author={Pankaj Kumar and Subhankar Mishra},
      year={2025},
      eprint={2505.18658},
      archivePrefix={arXiv},
      primaryClass={cs.CL},
      url={https://arxiv.org/abs/2505.18658}, 
}

@misc{feng2021surveydataaugmentationapproaches,
      title={A Survey of Data Augmentation Approaches for NLP}, 
      author={Steven Y. Feng and Varun Gangal and Jason Wei and Sarath Chandar and Soroush Vosoughi and Teruko Mitamura and Eduard Hovy},
      year={2021},
      eprint={2105.03075},
      archivePrefix={arXiv},
      primaryClass={cs.CL},
      url={https://arxiv.org/abs/2105.03075}, 
}

@misc{zhou2024surveydataaugmentationlarge,
      title={A Survey on Data Augmentation in Large Model Era}, 
      author={Yue Zhou and Chenlu Guo and Xu Wang and Yi Chang and Yuan Wu},
      year={2024},
      eprint={2401.15422},
      archivePrefix={arXiv},
      primaryClass={cs.LG},
      url={https://arxiv.org/abs/2401.15422}, 
}

@misc{chen2023fireactlanguageagentfinetuning,
      title={FireAct: Toward Language Agent Fine-tuning}, 
      author={Baian Chen and Chang Shu and Ehsan Shareghi and Nigel Collier and Karthik Narasimhan and Shunyu Yao},
      year={2023},
      eprint={2310.05915},
      archivePrefix={arXiv},
      primaryClass={cs.CL},
      url={https://arxiv.org/abs/2310.05915}, 
}

@inproceedings{zeng-etal-2024-agenttuning,
    title = "{A}gent{T}uning: Enabling Generalized Agent Abilities for {LLM}s",
    author = "Zeng, Aohan  and
      Liu, Mingdao  and
      Lu, Rui  and
      Wang, Bowen  and
      Liu, Xiao  and
      Dong, Yuxiao  and
      Tang, Jie",
    editor = "Ku, Lun-Wei  and
      Martins, Andre  and
      Srikumar, Vivek",
    booktitle = "Findings of the Association for Computational Linguistics: ACL 2024",
    month = aug,
    year = "2024",
    address = "Bangkok, Thailand",
    publisher = "Association for Computational Linguistics",
    url = "https://aclanthology.org/2024.findings-acl.181/",
    doi = "10.18653/v1/2024.findings-acl.181",
    pages = "3053--3077"
}

@inproceedings{chen-etal-2024-agent,
    title = "Agent-{FLAN}: Designing Data and Methods of Effective Agent Tuning for Large Language Models",
    author = "Chen, Zehui  and
      Liu, Kuikun  and
      Wang, Qiuchen  and
      Zhang, Wenwei  and
      Liu, Jiangning  and
      Lin, Dahua  and
      Chen, Kai  and
      Zhao, Feng",
    editor = "Ku, Lun-Wei  and
      Martins, Andre  and
      Srikumar, Vivek",
    booktitle = "Findings of the Association for Computational Linguistics: ACL 2024",
    month = aug,
    year = "2024",
    address = "Bangkok, Thailand",
    publisher = "Association for Computational Linguistics",
    url = "https://aclanthology.org/2024.findings-acl.557/",
    doi = "10.18653/v1/2024.findings-acl.557",
    pages = "9354--9366"
}

@misc{acikgoz2025singlemodelmastermultiturn,
      title={Can a Single Model Master Both Multi-turn Conversations and Tool Use? CoALM: A Unified Conversational Agentic Language Model}, 
      author={Emre Can Acikgoz and Jeremiah Greer and Akul Datta and Ze Yang and William Zeng and Oussama Elachqar and Emmanouil Koukoumidis and Dilek Hakkani-Tür and Gokhan Tur},
      year={2025},
      eprint={2502.08820},
      archivePrefix={arXiv},
      primaryClass={cs.AI},
      url={https://arxiv.org/abs/2502.08820}, 
}

@misc{wei2025autotirautonomoustoolsintegrated,
      title={AutoTIR: Autonomous Tools Integrated Reasoning via Reinforcement Learning}, 
      author={Yifan Wei and Xiaoyan Yu and Yixuan Weng and Tengfei Pan and Angsheng Li and Li Du},
      year={2025},
      eprint={2507.21836},
      archivePrefix={arXiv},
      primaryClass={cs.CL},
      url={https://arxiv.org/abs/2507.21836}, 
}

@misc{wu2025vtoolr1vlmslearnthink,
      title={VTool-R1: VLMs Learn to Think with Images via Reinforcement Learning on Multimodal Tool Use}, 
      author={Mingyuan Wu and Jingcheng Yang and Jize Jiang and Meitang Li and Kaizhuo Yan and Hanchao Yu and Minjia Zhang and Chengxiang Zhai and Klara Nahrstedt},
      year={2025},
      eprint={2505.19255},
      archivePrefix={arXiv},
      primaryClass={cs.LG},
      url={https://arxiv.org/abs/2505.19255}, 
}

@inproceedings{NEURIPS2024_e4c61f57,
 author = {Patil, Shishir G. and Zhang, Tianjun and Wang, Xin and Gonzalez, Joseph E.},
 booktitle = {Advances in Neural Information Processing Systems},
 doi = {10.52202/079017-4020},
 editor = {A. Globerson and L. Mackey and D. Belgrave and A. Fan and U. Paquet and J. Tomczak and C. Zhang},
 pages = {126544--126565},
 publisher = {Curran Associates, Inc.},
 title = {Gorilla: Large Language Model Connected with Massive APIs},
 url = {https://proceedings.neurips.cc/paper_files/paper/2024/file/e4c61f578ff07830f5c37378dd3ecb0d-Paper-Conference.pdf},
 volume = {37},
 year = {2024}
}

@inproceedings{
qin2024toolllm,
title={Tool{LLM}: Facilitating Large Language Models to Master 16000+ Real-world {API}s},
author={Yujia Qin and Shihao Liang and Yining Ye and Kunlun Zhu and Lan Yan and Yaxi Lu and Yankai Lin and Xin Cong and Xiangru Tang and Bill Qian and Sihan Zhao and Lauren Hong and Runchu Tian and Ruobing Xie and Jie Zhou and Mark Gerstein and dahai li and Zhiyuan Liu and Maosong Sun},
booktitle={The Twelfth International Conference on Learning Representations},
year={2024},
url={https://openreview.net/forum?id=dHng2O0Jjr}
}

@inproceedings{
shi2023language,
title={Language models are multilingual chain-of-thought reasoners},
author={Freda Shi and Mirac Suzgun and Markus Freitag and Xuezhi Wang and Suraj Srivats and Soroush Vosoughi and Hyung Won Chung and Yi Tay and Sebastian Ruder and Denny Zhou and Dipanjan Das and Jason Wei},
booktitle={The Eleventh International Conference on Learning Representations },
year={2023},
url={https://openreview.net/forum?id=fR3wGCk-IXp}
}

@inproceedings{moradshahi-etal-2023-x,
    title = "{X}-{R}i{SAWOZ}: High-Quality End-to-End Multilingual Dialogue Datasets and Few-shot Agents",
    author = "Moradshahi, Mehrad  and
      Shen, Tianhao  and
      Bali, Kalika  and
      Choudhury, Monojit  and
      de Chalendar, Gael  and
      Goel, Anmol  and
      Kim, Sungkyun  and
      Kodali, Prashant  and
      Kumaraguru, Ponnurangam  and
      Semmar, Nasredine  and
      Semnani, Sina  and
      Seo, Jiwon  and
      Seshadri, Vivek  and
      Shrivastava, Manish  and
      Sun, Michael  and
      Yadavalli, Aditya  and
      You, Chaobin  and
      Xiong, Deyi  and
      Lam, Monica",
    editor = "Rogers, Anna  and
      Boyd-Graber, Jordan  and
      Okazaki, Naoaki",
    booktitle = "Findings of the Association for Computational Linguistics: ACL 2023",
    month = jul,
    year = "2023",
    address = "Toronto, Canada",
    publisher = "Association for Computational Linguistics",
    url = "https://aclanthology.org/2023.findings-acl.174/",
    doi = "10.18653/v1/2023.findings-acl.174",
    pages = "2773--2794"
}

@misc{parisi2022talmtoolaugmentedlanguage,
      title={TALM: Tool Augmented Language Models}, 
      author={Aaron Parisi and Yao Zhao and Noah Fiedel},
      year={2022},
      eprint={2205.12255},
      archivePrefix={arXiv},
      primaryClass={cs.CL},
      url={https://arxiv.org/abs/2205.12255}, 
}

@article{schick2023toolformer,
  title={Toolformer: Language models can teach themselves to use tools},
  author={Schick, Timo and Dwivedi-Yu, Jane and Dess{\`\i}, Roberto and Raileanu, Roberta and Lomeli, Maria and Hambro, Eric and Zettlemoyer, Luke and Cancedda, Nicola and Scialom, Thomas},
  journal={Advances in Neural Information Processing Systems},
  volume={36},
  pages={68539--68551},
  year={2023}
}

@article{tang2023toolalpaca,
  title={Toolalpaca: Generalized tool learning for language models with 3000 simulated cases, 2023},
  author={Tang, Qiaoyu and Deng, Ziliang and Lin, Hongyu and Han, Xianpei and Liang, Qiao and Cao, Boxi and Sun, Le},
  journal={URL https://arxiv. org/abs/2306.05301},
  year={2023}
}

@inproceedings{chen2024t,
  title={T-eval: Evaluating the tool utilization capability of large language models step by step},
  author={Chen, Zehui and Du, Weihua and Zhang, Wenwei and Liu, Kuikun and Liu, Jiangning and Zheng, Miao and Zhuo, Jingming and Zhang, Songyang and Lin, Dahua and Chen, Kai and others},
  booktitle={Proceedings of the 62nd Annual Meeting of the Association for Computational Linguistics (Volume 1: Long Papers)},
  pages={9510--9529},
  year={2024}
}

@article{wang2023mint,
  title={Mint: Evaluating llms in multi-turn interaction with tools and language feedback},
  author={Wang, Xingyao and Wang, Zihan and Liu, Jiateng and Chen, Yangyi and Yuan, Lifan and Peng, Hao and Ji, Heng},
  journal={arXiv preprint arXiv:2309.10691},
  year={2023}
}

\newpage
\appendix
\section{Appendix}
\subsection{Code and Dataset}
\label{sec:code}
All code and datasets used in this paper are available at \url{https://anonymous.4open.science/r/multilingual_robustness_tool_calling-CA44}.

\subsection{Detailed Error Taxonomy}
\label{app:sec:error}
To characterize multilingual tool-calling failures beyond binary correctness, we organize errors according to their impact on execution rather than semantic adequacy alone.
As illustrated in Figure~\ref{fig:error-tax}, error categories are ordered by severity, reflecting whether a generated tool call can be parsed, executed, or reliably recovered in a realistic tool-calling system.

At the most severe end are \textbf{syntax errors}, where malformed outputs cannot be parsed into valid function invocations.
These failures prevent any execution attempt and reflect violations of the output format rather than language understanding.
\textbf{Function-level errors} are syntactically valid but violate the tool schema, such as incorrect function names or missing parameters.
While closer to execution, they still result in non-executable tool calls.

A central focus of this work is \emph{parameter value language mismatch}, which we further decompose by semantic correctness. 
(a) \textbf{Language Mismatch + Wrong Value:}
In the most severe cases, parameter values are expressed in a non-English language and are semantically unrelated to the ground truth.
(b) \textbf{Language Mismatch + Relevant but Incorrect:} Less severe variants preserve partial relevance but do not precisely match the intended meaning. They typically arise when models mirror the user’s language while producing under-specified values, reflecting combined language and semantic imprecision.
(c) \textbf{Language Mismatch + Same Meaning:} Another less severe variant shows non-English parameter values that are conceptually related but do not precisely match the intended meaning. They typically arise when models mirror the user’s language while producing under-specified values, reflecting combined language and semantic imprecision.
Together, these errors isolate failures at the language–execution interface, as intent understanding and tool selection have already succeeded.

Finally, we consider parameter value errors without language mismatch, ranging from semantically incorrect and irrelevant values (\textbf{Wrong Value}) to semantically relevant but incorrect values (\textbf{Relevant but Incorrect}), values that are semantically the same but differ in surface form (\textbf{Exactly Same Meaning}), thus fail strict BFCL surface-form matching.

\subsection{Details of Inference-Time Mitigation Strategies}
\label{sec:mitigation-details}

\subsubsection{Prompt-Level Instruction (PT)}

Prompt-Level Instruction (PT) fully translates the user query into the target language and explicitly instructs the model to output parameter values in English.
This strategy probes whether models can follow execution-level conventions through natural-language instructions alone, without altering the input query or post-processing the output.

\subsubsection{Pre-Translation (PRE)}
Pre-Translation of User Queries (PRE) translates non-English user queries into English before tool calling.
To avoid confounding the effect of translation quality with model capability, the same language model is used for both translation and tool calling.
PRE removes multilingual input entirely from the tool-calling step, serving as an upper bound on what can be achieved through input normalization.

\subsubsection{Post-Translation (POST)}
Post-Translation of Parameter Values (POST) translates generated parameter values into English after tool calling but before execution.
This strategy directly targets parameter value language mismatch while preserving the model’s original tool selection and argument structure.
As with PRE, translation is performed by the same model to control for language proficiency.

\subsection{Manual Verification Process}
\label{sec:verification}
We have group members who speak Chinese, Hindi, and Igbo.

For each entry of the fully translated dataset, we verify that the translated user queries have the same meaning as the original ones, but we do not enforce the wordings of the queries so that a tool-calling LLM intelligent enough always has a chance to hit the ground truth. For example, ``queen-size bed" may not have a strict equivalence in many languages; therefore, we may judge any translation that is semantically equivalent to ``a large bed" as appropriate. As a consequence, we do not expect the tool-calling LLM to recover exactly ``queen-sized bed", but if the LLM fails to do so, we still count it as a failure to meet the ground truth.

For the partially translated dataset, we verify both the semantic invariance and the property that keywords acting as parameter values in the ground truth are kept in English. It is noteworthy that this kind of keyword preserving may not stop LLMs from passing in parameter values that are not in English. For example, if only ``A" in ``Company A" appears in the ground truth parameter values, we do not also keep the word ``Company" in English, although LLMs may pass in ``Company A" in the user query's language, causing a language mismatch. 

\subsection{Sample Tool Call Generation Prompt}
\label{sec:tool-call-generation-prompt}

We use BFCL's prompt for tool generation, and add slight modifications to it to adapt to the properties of different models (for example, GPT-5 uses ``developer" message instead of ``system" message). A sample prompt can be found below. Note that the ``IMPORTANT: Pass all parameter values in English" prompt only appears in the \textbf{PT} setting, as illustrated in Figure~\ref{fig:tool-benchmark}.

\begin{promptbox}
\textbf{System Prompt:}

You are an expert in composing functions. 
You are given a question and a set of possible functions. 
Based on the question, you will need to make one or more function/tool calls to achieve the purpose. 
If none of the functions can be used, point it out. 
If the given question lacks the parameters required by the function, also point it out.

You should ONLY return function calls in your response. 
You MUST NOT include any other text, explanations, or direct answers. 
If you decide to invoke any function(s), you MUST use the provided tools. 
Do NOT attempt to answer the question directly without using the available functions.

[IMPORTANT: Pass all parameter values in English.]

\textbf{User Prompt:}

[user query]

\end{promptbox}

\subsection{Fully Translated Dataset Generation Prompt}
\label{sec:unverified_ft_prompt}
\begin{promptbox}
\textbf{Prompt:}

Translate the following English question to [target\_language]. Provide a natural, fluent translation that maintains the meaning and intent of the original question.
\end{promptbox}

\subsection{Partially Translated Dataset Generation Prompt}
\label{sec:unverified_pt_prompt}
A sample prompt can be found below.
Note that the keywords are extracted from the parameter values in the ground truth.

\begin{promptbox}
\textbf{Prompt:}

Translate the following English question to [target\_language]. Keep the keywords listed below unchanged (do not translate them).

Question:
[question\_content]

Keywords to preserve (keep in English):
[keywords\_str]

Provide only the [target\_language] translation.
\end{promptbox}

\subsection{Paraphrased Dataset Generation Prompt}
\label{sec:paraphrase_prompt}
\begin{promptbox}
\textbf{Prompt:}

You are a helpful assistant helping rephrase user requests, while accurately preserving their meaning, including numbers and names if they exist. Do not answer the requirement; just produce another one that is identical in meaning but is phrased differently. Produce ONLY the rephrased requirement, without further thoughts or explanations.
\end{promptbox}

\subsection{Synonym Dataset Generation Prompt}
\label{sec:synonym_prompt}
\begin{promptbox}
\textbf{Prompt:}

You are a helpful assistant that replaces words with synonyms of similar meaning while maintaining semantic correctness. Your task is to process word by word and replace each word with a synonym if possible.

IMPORTANT RULES:
1. Replace words with appropriate synonyms
2. Maintain the semantic meaning and grammatical structure
3. Do NOT perform general paraphrasing, only synonym replacement
4. Process word by word, not phrase by phrase
5. If a word has no suitable synonym or is a proper noun, keep it unchanged

Produce ONLY the modified text with synonyms, without further thoughts or explanations. Consider the example below:

USER: Can I find the dimensions and properties of a triangle, if it is known that its three sides are 5 units, 4 units and 3 units long?

ASSISTANT: Can I discover the measurements and characteristics of a triangle, if it is known that its three sides are 5 units, 4 units and 3 units long?
\end{promptbox}

\subsection{Tool Calling IO Protocol for Tested Model Families}
\label{sec:official}
We do not use the official BFCL repository's implementation of tool calling protocol handling, but implement it ourselves to stick to the latest official documentation for each model and make the code base clean and extensible. The following are the documentations we reference to implement the tool, calling format conversion, and output parsing. The detailed implementation can be viewed at Appendix \ref{sec:code}.
\paragraph{GPT-5 Family}
The official documentation for the GPT-5 tool calling IO protocol can be found at \url{https://platform.openai.com/docs/guides/function-calling?strict-mode=enabled#defining-functions}.
\paragraph{DeepSeek V3.2}
The official documentation for the DeepSeek V3.2 tool calling IO protocol can be found at
\url{https://api-docs.deepseek.com/guides/function_calling}.
\paragraph{Qwen 3 Family}
The official documentation for the Qwen 3 tool calling IO protocol can be found at \url{https://qwen.readthedocs.io/en/latest/framework/function_call.html}.
\paragraph{Llama 3.1 Family}
The official documentation for the Llama 3.1 tool calling IO protocol can be found at \url{https://huggingface.co/meta-llama/Llama-3.1-70B-Instruct#tool-use-with-transformers}.
\paragraph{Granite 4 Family}
The official documentation for Granite 4 tool calling IO protocol can be found at \url{https://huggingface.co/ibm-granite/granite-4.0-micro}.

\section{Additional Results}

\subsection{Case Study of Translation-Induced Semantic Drift.}
\label{app:sec:case}
\begin{figure}[h]
  \centering
  \includegraphics[width=\columnwidth]{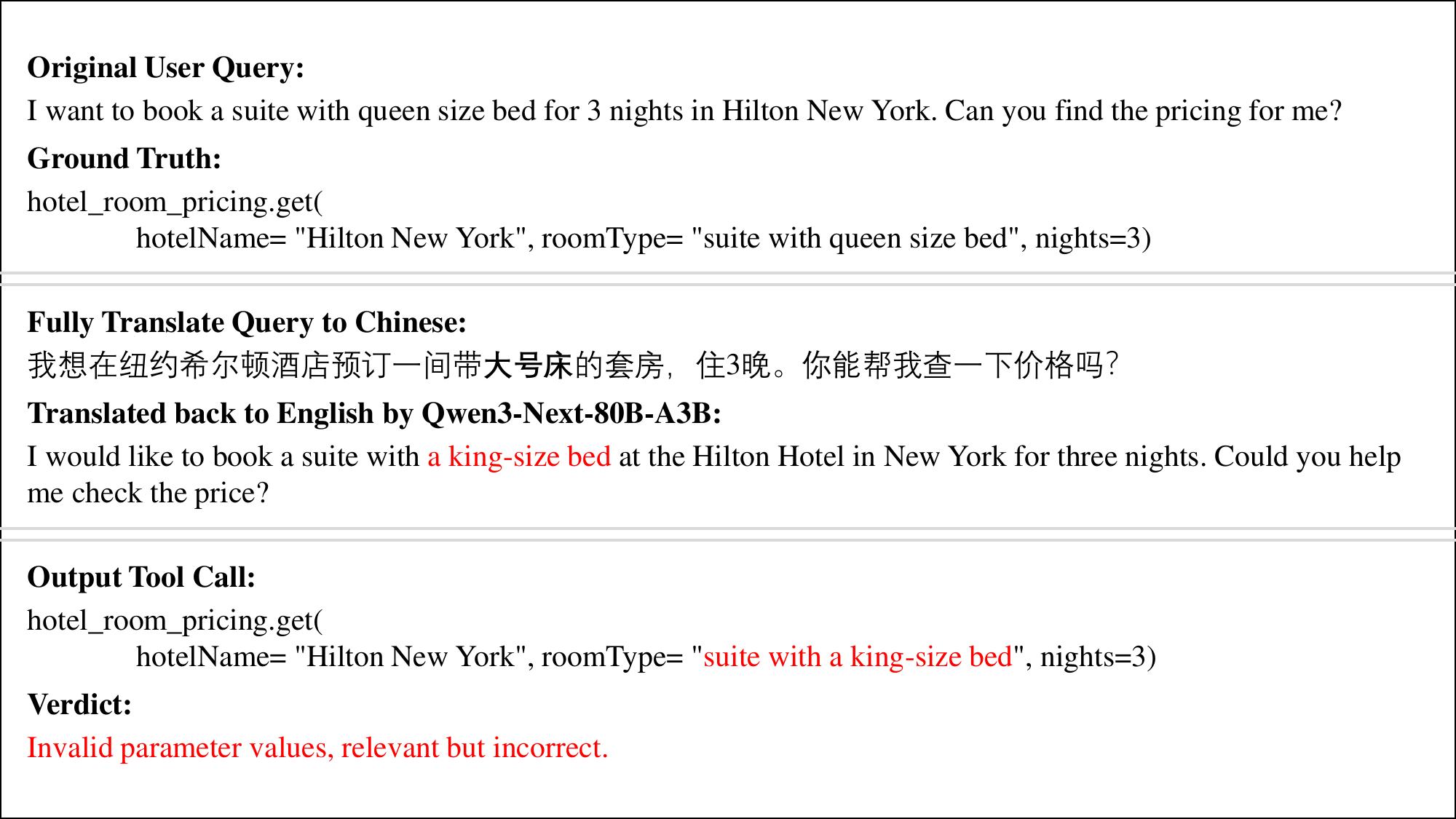}
\caption{ {Case study: semantic drift introduced by query-level translation.}
An example where pre-translating a user query alters the realized parameter surface forms, leading to an execution failure despite correct intent understanding.}

  \label{fig:pre-translation}
\end{figure}

Figure~\ref{fig:pre-translation} presents a representative example of pre-translation (PRE), where translating the user query alters parameter surface forms through normalization or synonym substitution.
Although the translated query preserves the original intent, the resulting tool call fails under strict surface-form matching, illustrating why PRE cannot fully recover English-level performance.

\subsection{Effects of Model Scale and Architecture}
\label{sec:effects-of-model-scale}
\begin{figure}[t]
  \centering
  \subfigure[\textbf{Fully translated queries (FT).}]{
    \includegraphics[width=0.49\textwidth]{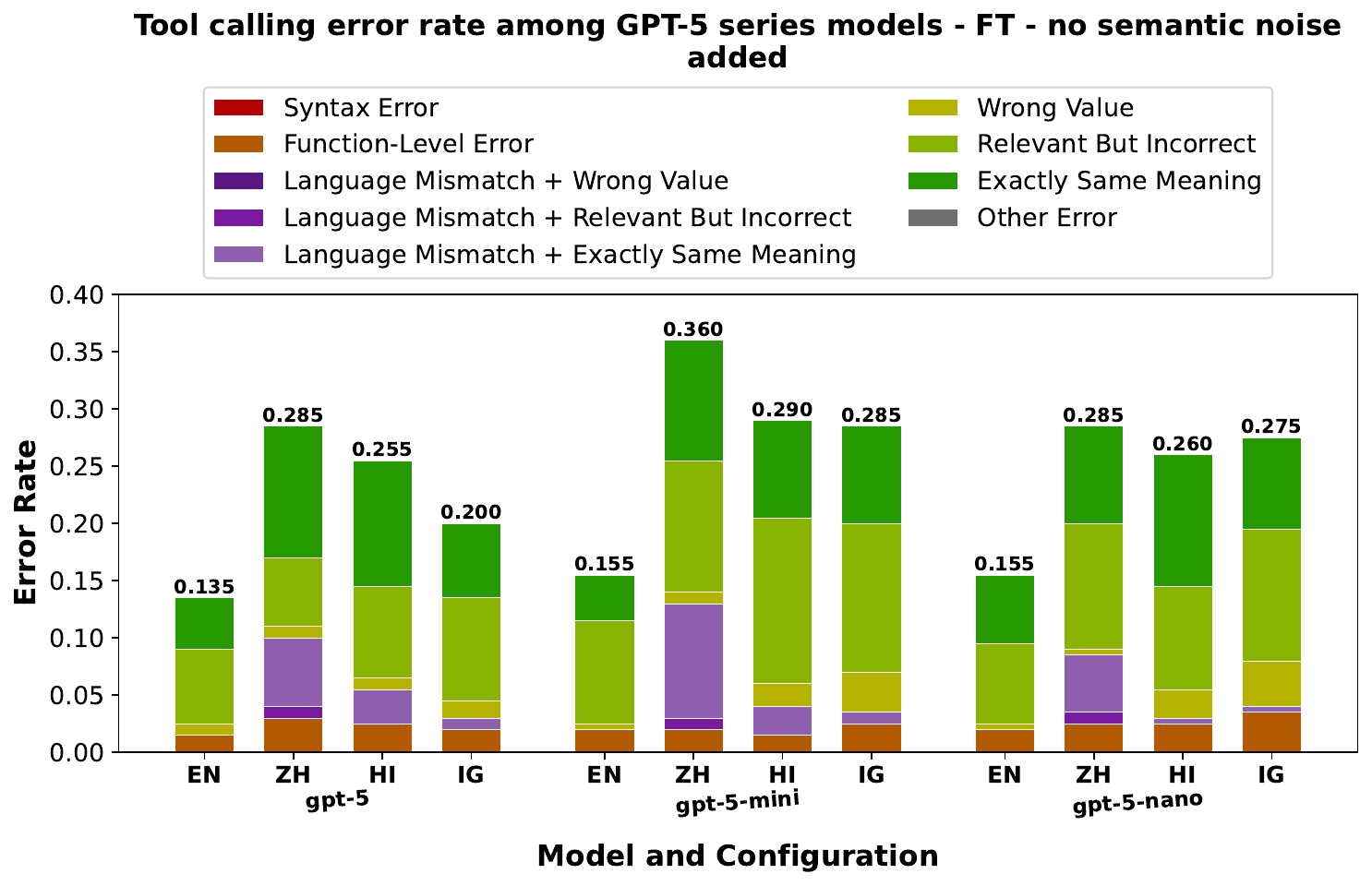}
    \label{fig:by_model_Gpt5_ft}
  }
  \hfill
  \subfigure[\textbf{Partially translated queries (PAR).}]{
    \includegraphics[width=0.49\textwidth]{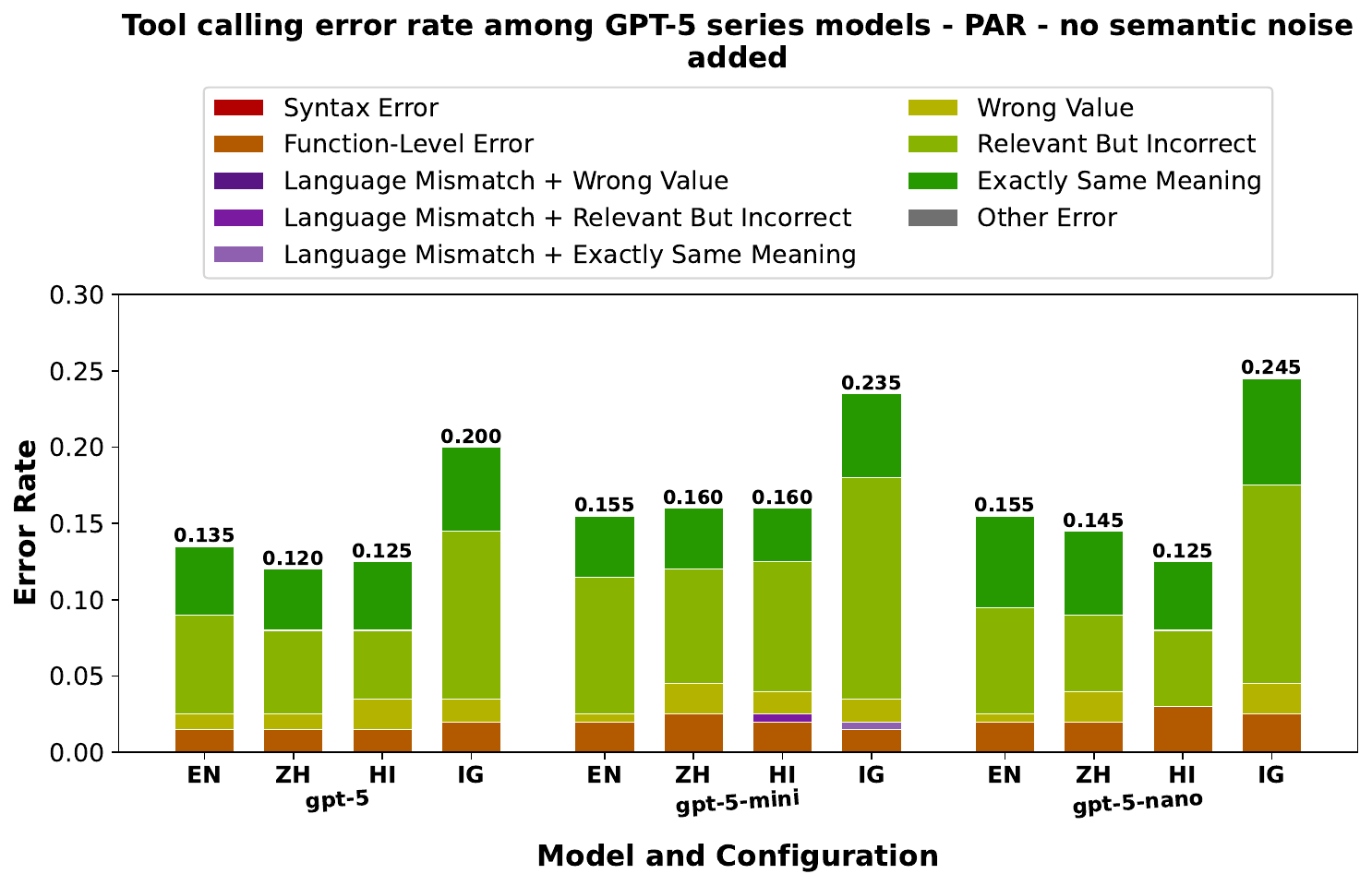}
    \label{fig:by_model_Gpt5_par}
  }
  \hfill
  \subfigure[\textbf{Fully translated with English-parameter prompting (PT).}]{
    \includegraphics[width=0.49\textwidth]{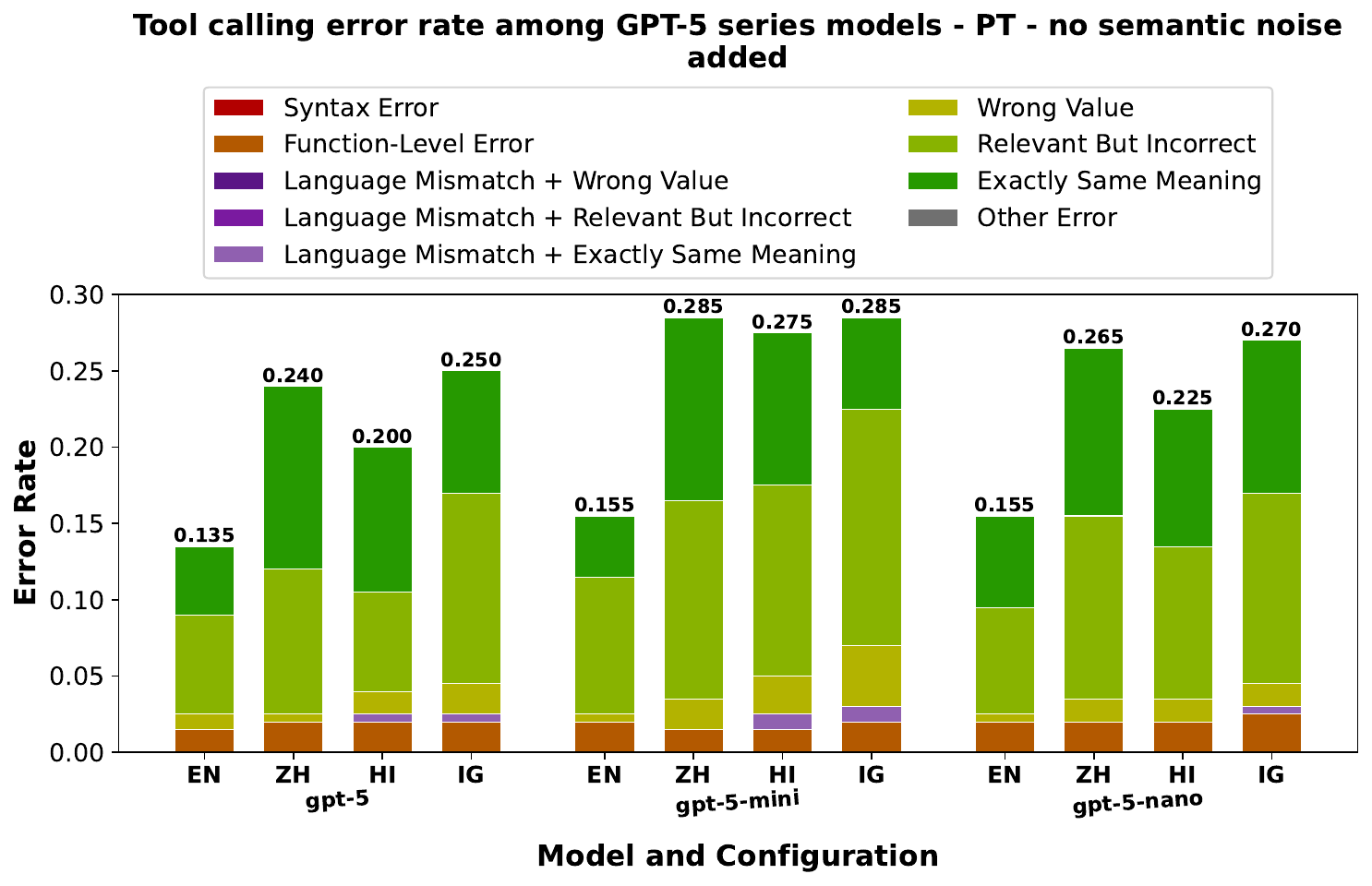}
    \label{fig:by_model_Gpt5_pt}
  }

  \caption{\textbf{GPT-5 series behavior across query language compositions and inference-time prompting.}
  We compare GPT-5, GPT-5 mini, and GPT-5 nano under fully translated (FT), partially translated (PAR), and fully translated queries with explicit English-parameter prompting (PT).
  Across model sizes, GPT-5 variants exhibit low semantic error rates, indicating strong multilingual intent understanding.
  Errors in the FT setting are primarily driven by parameter value language mismatch, which is substantially reduced in PAR and PT, demonstrating that execution-level language conventions, rather than semantic reasoning, dominate GPT-5 failures under multilingual tool calling.}
  \label{fig:gpt5_by_setting}
\end{figure}

Within the GPT-5 family (Figures~\ref{fig:by_model_Gpt5_ft}, \ref{fig:by_model_Gpt5_par}, \ref{fig:by_model_Gpt5_pt}), performance is consistent across scales, with the standard model slightly outperforming distilled variants. This indicates that tool-calling accuracy depends on the model's style at comprehending the user queries and its understanding of the best tool-calling convention to determine things like how many functions to call, what parameter values to choose, etc, under limited constraint or instruction, to align with the input convention of the downstream task.

\begin{figure}[t]
  \centering
  \subfigure[\textbf{Fully translated queries (FT).}]{
    \includegraphics[width=0.49\textwidth]{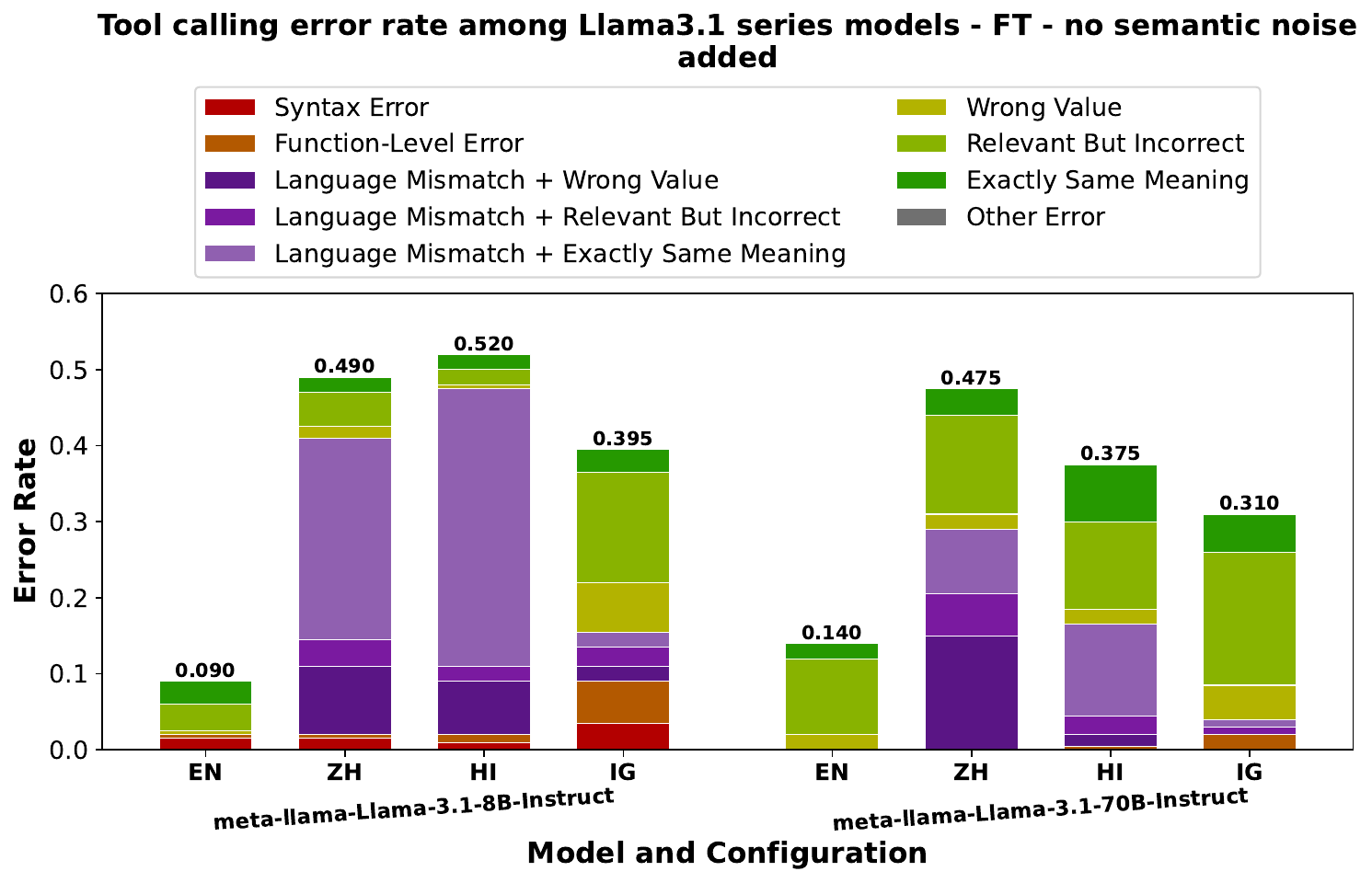}
    \label{fig:by_model_llama3.1_ft}
  }
  \hfill
  \subfigure[\textbf{Partially translated queries (PAR).}]{
    \includegraphics[width=0.49\textwidth]{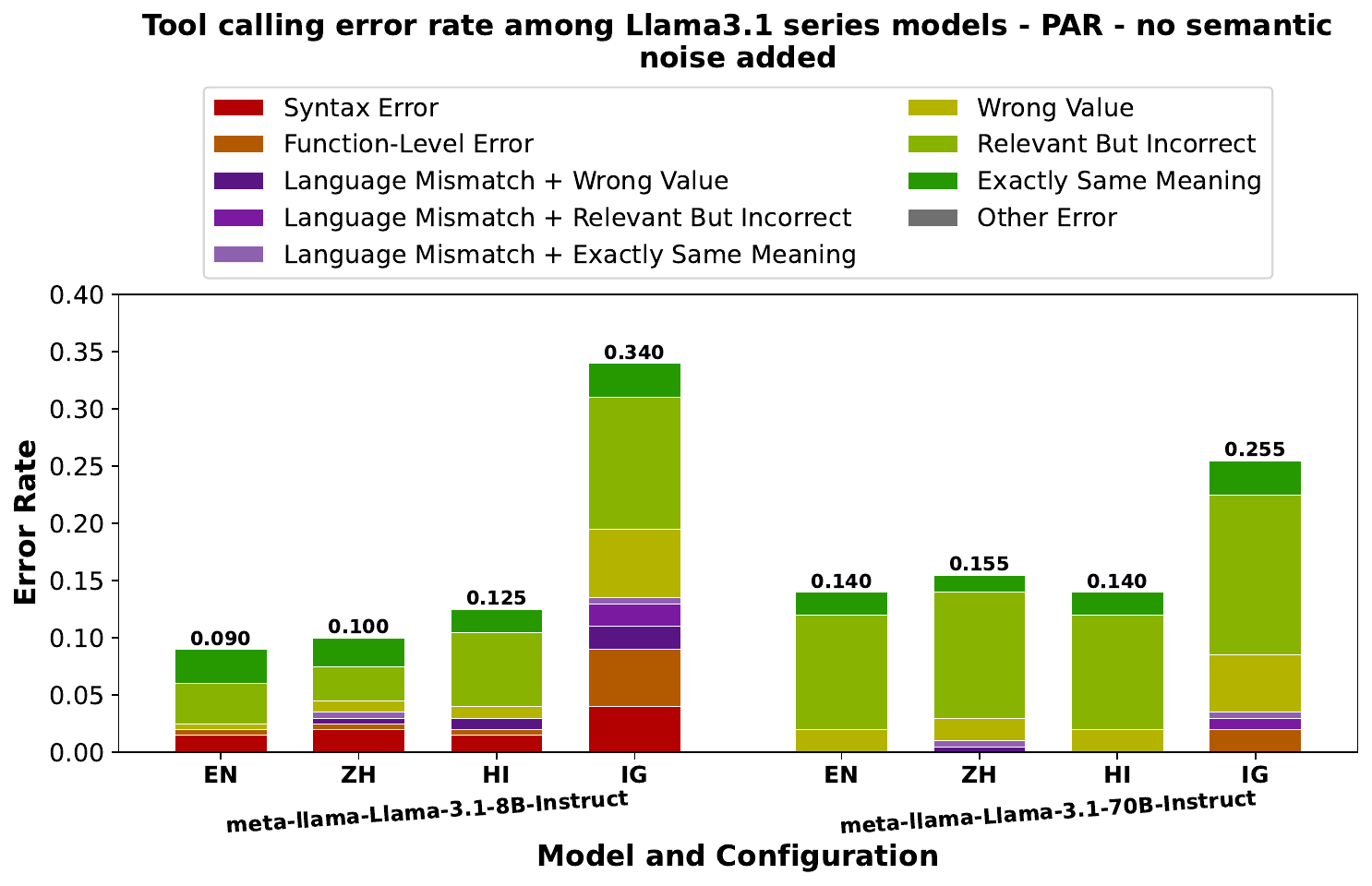}
    \label{fig:by_model_llama3.1_par}
  }
  \hfill
  \subfigure[\textbf{Fully translated with English-parameter prompting (PT).}]{
    \includegraphics[width=0.49\textwidth]{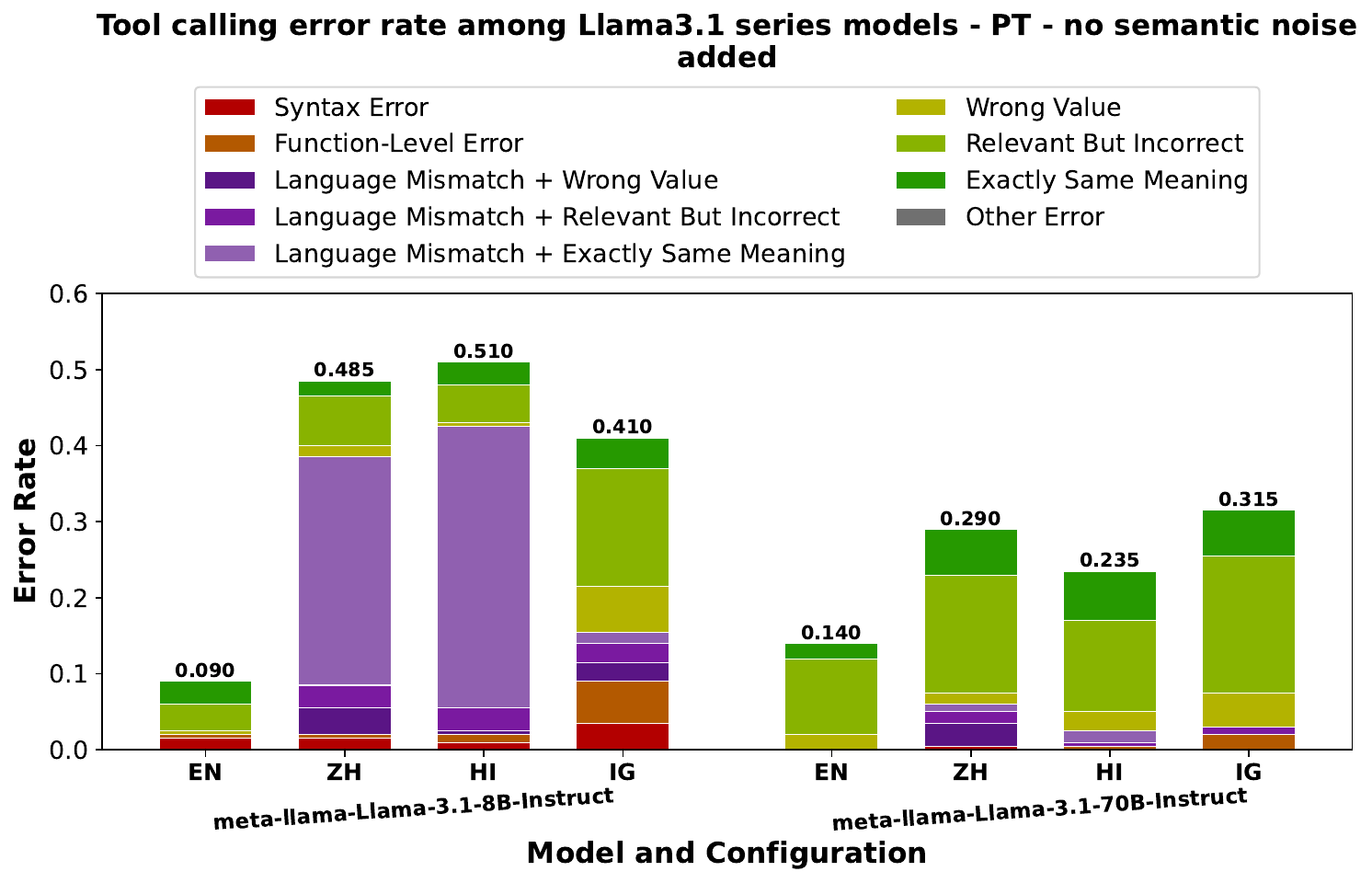}
    \label{fig:by_model_llama3.1_pt}
  }

  \caption{\textbf{Llama 3.1 series exhibits limited robustness to multilingual execution constraints.}
  We compare Llama~3.1--8B and Llama~3.1--70B under fully translated (FT), partially translated (PAR), and fully translated queries with explicit English-parameter prompting (PT).
  In contrast to GPT-5, Llama~3.1 shows substantial execution failures in the FT setting that combine parameter value language mismatch with semantic errors.
  While PAR and PT reduce some mismatch errors, a non-trivial portion of failures persists, indicating that multilingual degradation for Llama~3.1 reflects both execution-interface violations and reduced robustness in cross-lingual semantic interpretation.}
  \label{fig:llama31_by_setting}
\end{figure}

For Llama 3.1 (Figures~\ref{fig:by_model_llama3.1_ft}, \ref{fig:by_model_llama3.1_par}, \ref{fig:by_model_llama3.1_pt}), larger models show fewer syntax errors and stronger instruction following, including implicit awareness of English parameter conventions. Larger models also have a much better instruction following capability when being prompted to pass in parameter values in English, as shown in the PT experiment.
In short, for the Llama 3.1 family models, a large model size does improve the overall robustness under user queries of different languages. The improvement is composed of the stability of generating a valid tool call under confusion and uncertainty, a better awareness of the parameter value passing convention, and a better instruction following capability.

We also notice that in the partially translated experiment, the Llama 3.1 70B model has slightly worse performance than the Llama 3.1 8B model, although it has fewer syntax errors. After investigating the evaluation results, we find that for all numeric values, Llama 3.1 70B appears to only be able to output integers, while the ground truths are decimal numbers. We exclude the possibility of bugs in the parsing logic, and the framework's soundness is further backed up by the fact that Llama 3.1 8B uses the same framework while not having the same problem. This phenomenon remains unexplained.

\begin{figure}[t]
  \centering
  \subfigure[\textbf{Fully translated queries (FT).}]{
    \includegraphics[width=0.49\textwidth]{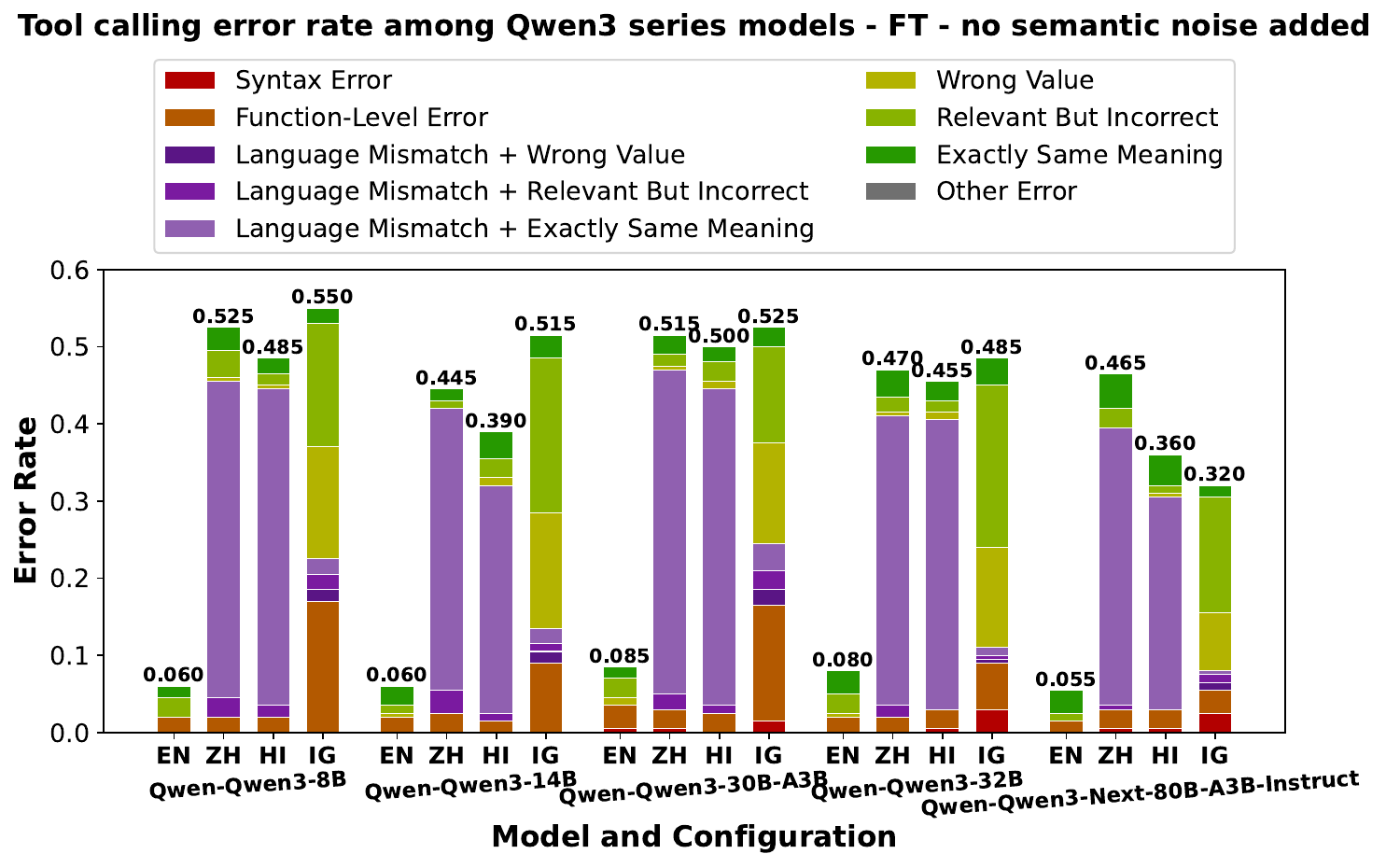}
    \label{fig:by_model_qwen3_ft}
  }
  \hfill
  \subfigure[\textbf{Partially translated queries (PAR).}]{
    \includegraphics[width=0.49\textwidth]{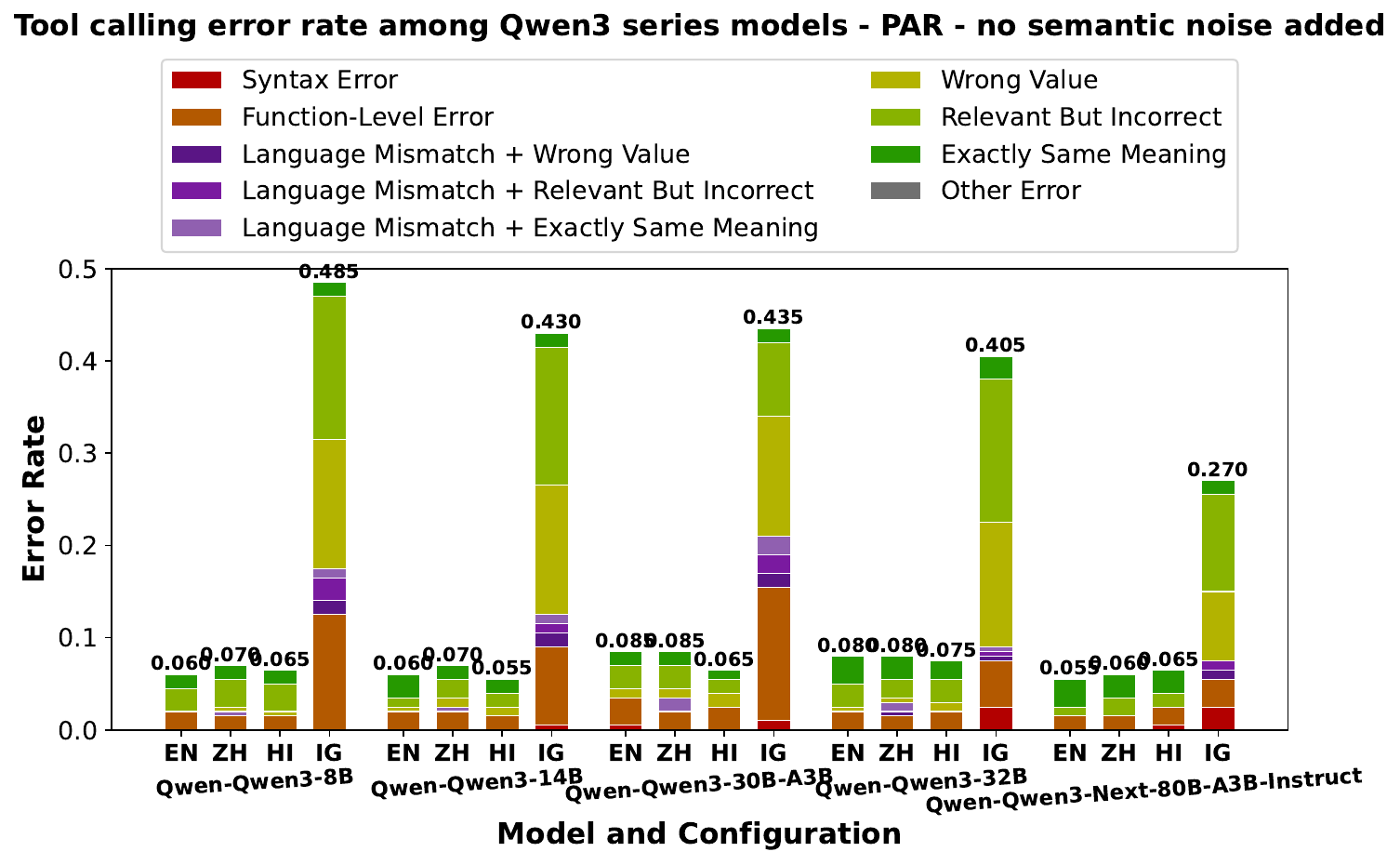}
    \label{fig:by_model_qwen3_par}
  }
  \hfill
  \subfigure[\textbf{Fully translated with English-parameter prompting (PT).}]{
    \includegraphics[width=0.49\textwidth]{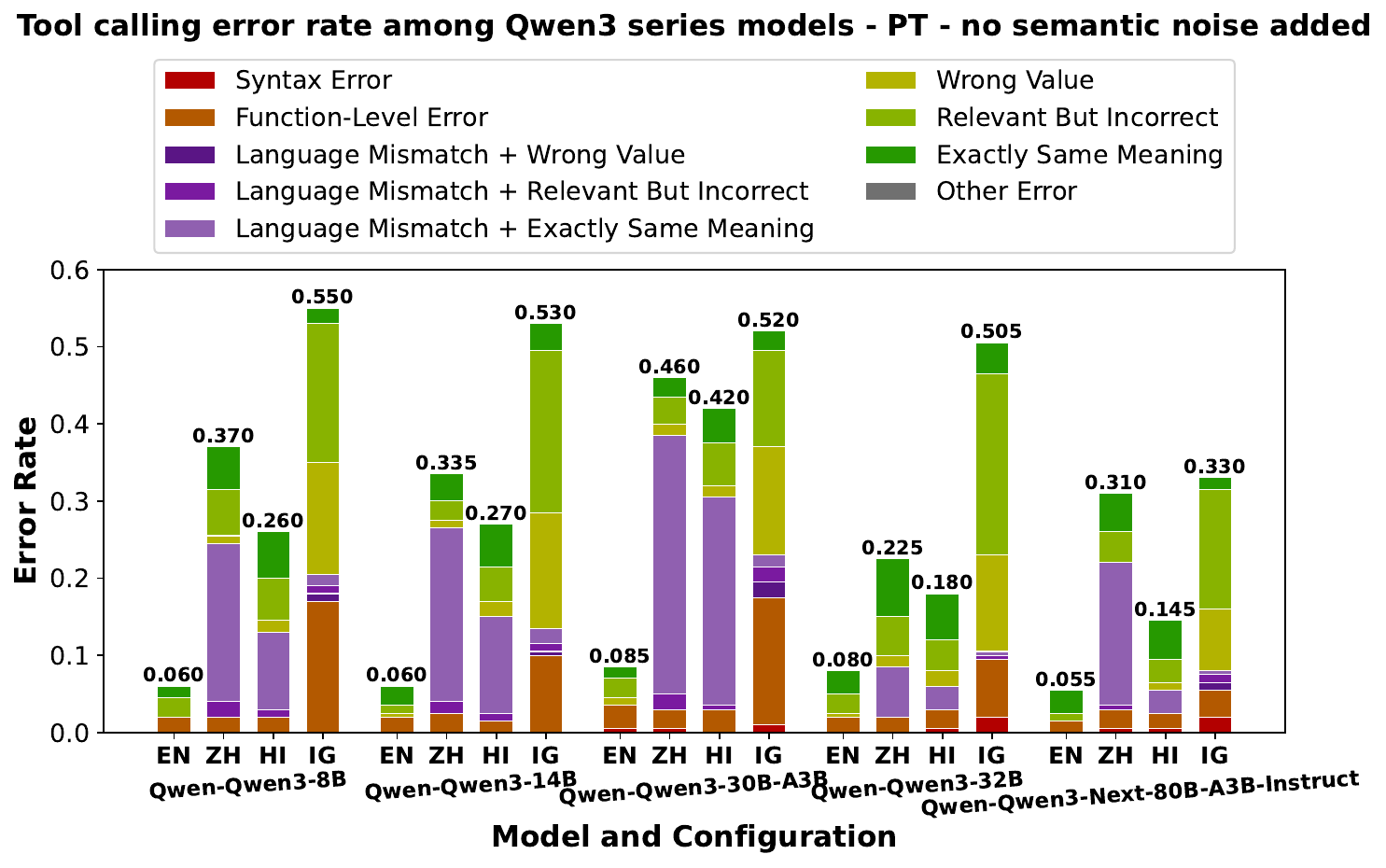}
    \label{fig:by_model_qwen3_pt}
  }

  \caption{\textbf{Qwen~3 models exhibit strong language alignment but inconsistent adherence to execution conventions.}
  We compare Qwen~3 models of different sizes under fully translated (FT), partially translated (PAR), and fully translated queries with explicit English-parameter prompting (PT).
  In the FT setting, Qwen~3 frequently preserves non-English parameter values, leading to prominent execution-level language mismatch errors.
  While PAR substantially reduces these errors by preserving English parameter strings, explicit prompting (PT) yields mixed improvements, indicating that strong language alignment can conflict with strict execution-interface requirements.}
  \label{fig:qwen3_by_setting}
\end{figure}

Qwen3 MoE models (Figures~\ref{fig:by_model_qwen3_ft}, \ref{fig:by_model_qwen3_par}, \ref{fig:by_model_qwen3_pt}) do not show monotonic gains with scale. Qwen3-30B-A3B underperforms Qwen3-14B, and Qwen3-Next-80B-A3B shows limited gains over Qwen3-32B. This is mainly because they are Mixture of Experts (MoE) models, where only a part of the neurons of the models are activated during inference, selected depending on the category of the task. This may cause the models to not have the expertise in both multilingual understanding and tool calling formation at the same time, thus having worse performance than smaller non-MoE models. 
Notably, Qwen3-Next-80B-A3B performs best on Igbo, showing that it has more knowledge about the Igbo language than the smaller models.

\begin{figure}[t]
  \centering
  \subfigure[\textbf{Fully translated queries (FT).}]{
    \includegraphics[width=0.49\textwidth]{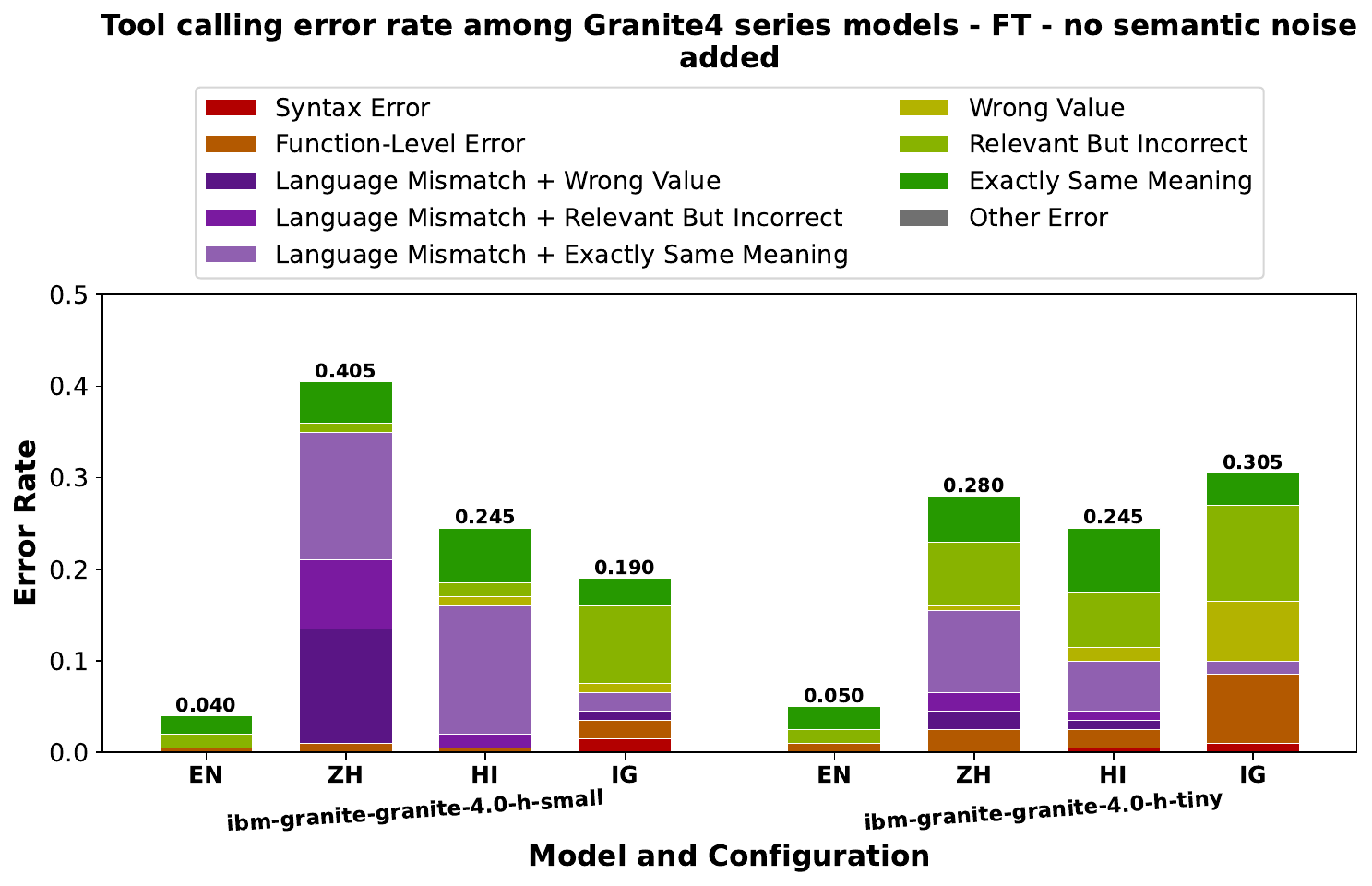}
    \label{fig:by_model_Granite4_ft}
  }
  \hfill
  \subfigure[\textbf{Partially translated queries (PAR).}]{
    \includegraphics[width=0.49\textwidth]{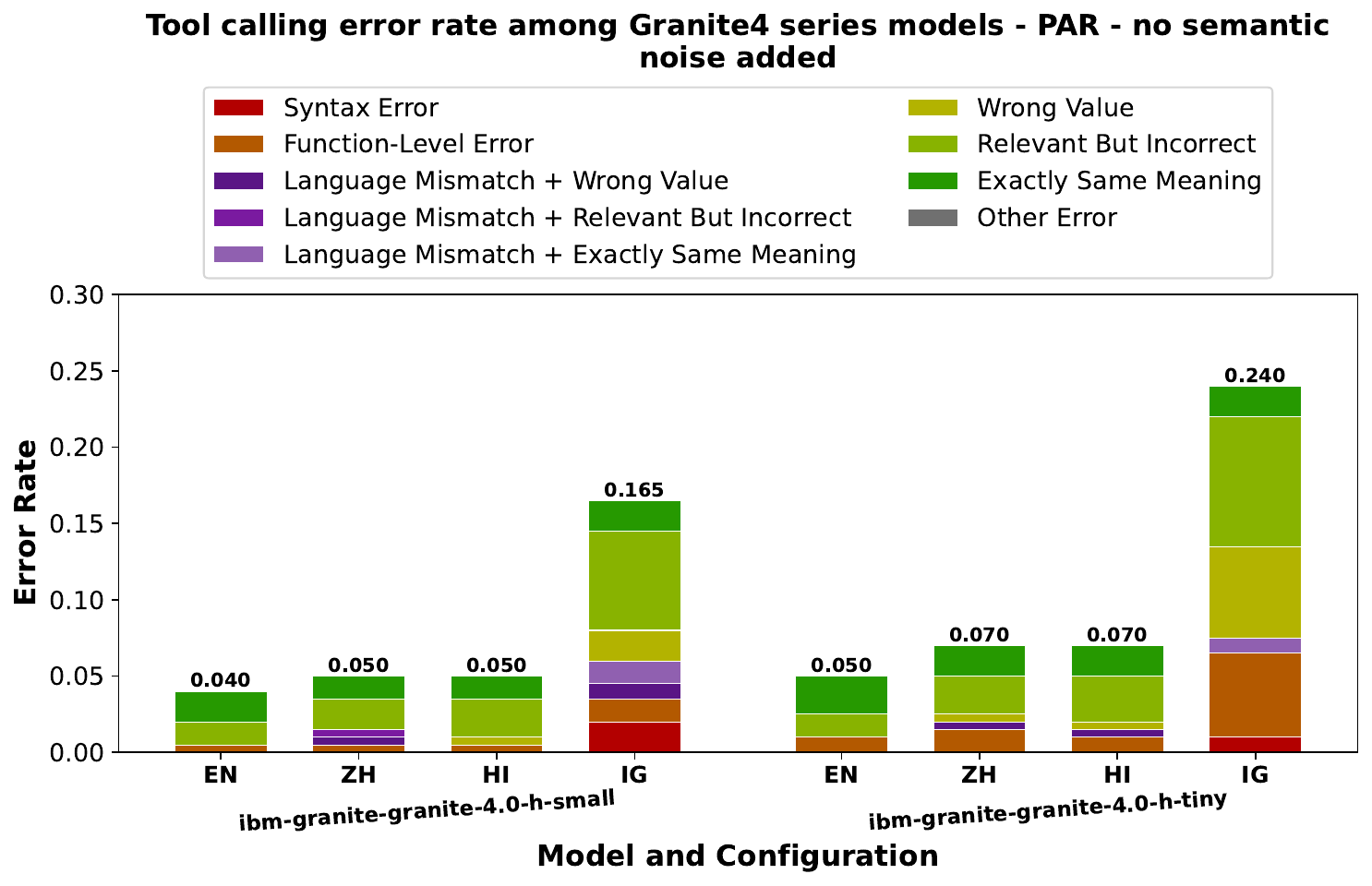}
    \label{fig:by_model_Granite4_par}
  }
  \hfill
  \subfigure[\textbf{Fully translated with English-parameter prompting (PT).}]{
    \includegraphics[width=0.49\textwidth]{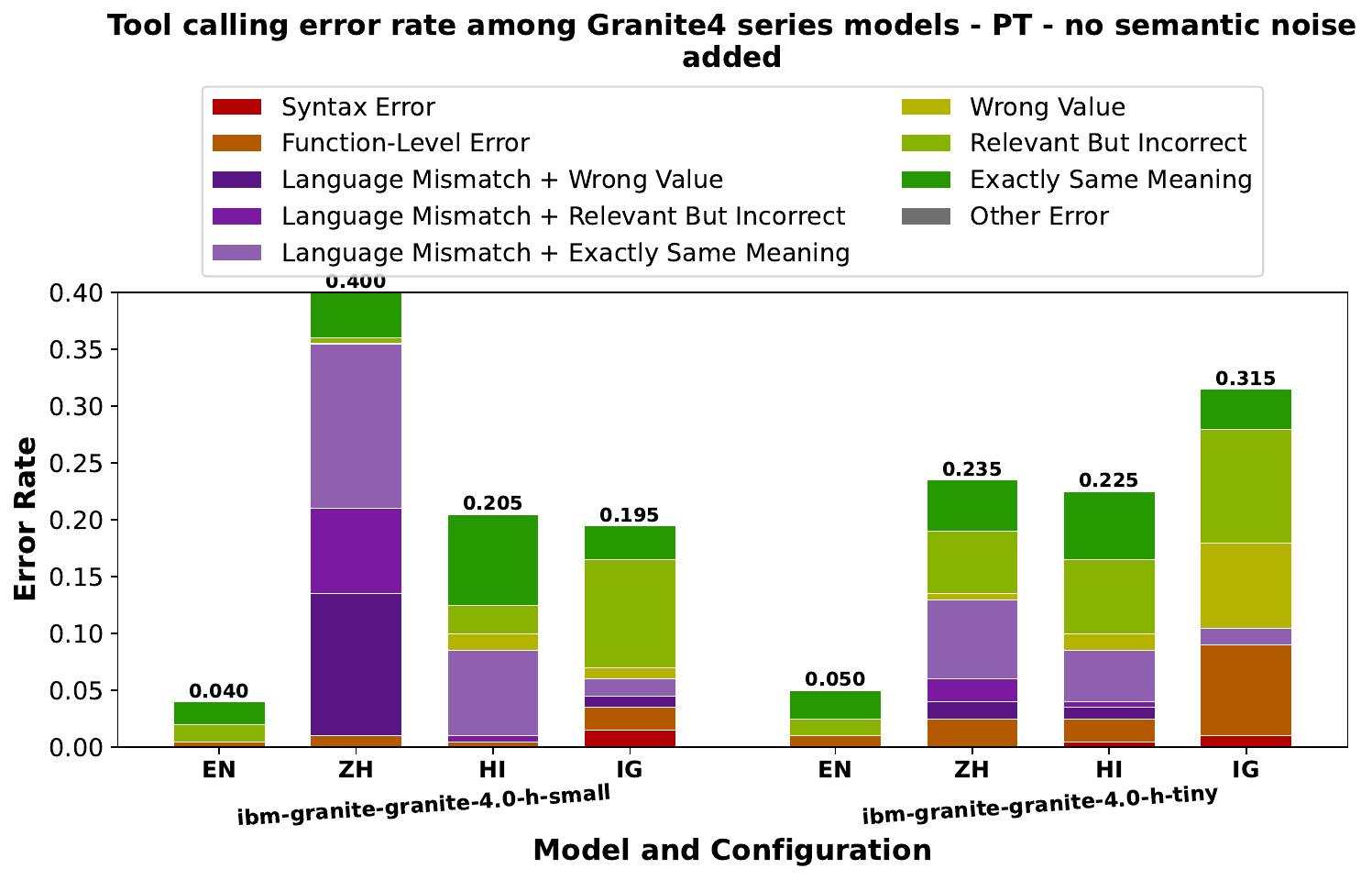}
    \label{fig:by_model_Granite4_pt}
  }

  \caption{\textbf{Granite~4 models show strong English tool-calling performance but limited gains from inference-time mitigation.}
  We evaluate Granite~4 models under fully translated (FT), partially translated (PAR), and fully translated queries with explicit English-parameter prompting (PT).
  While PAR reduces execution-level language mismatch by preserving English parameter strings, explicit prompting (PT) yields limited additional improvement.
  This suggests that Granite~4’s tool-calling behavior is less responsive to instruction-based mitigation, likely reflecting tighter coupling between tool execution patterns and training-time conventions.}
  \label{fig:granite4_by_setting}
\end{figure}

Granite 4 models (Figures~\ref{fig:by_model_Granite4_ft}, \ref{fig:by_model_Granite4_par}, and \ref{fig:by_model_Granite4_pt}) show an inverse trend: larger models perform worse on Chinese but better on Igbo.   This suggests that Granite~4’s tool-calling behavior is less responsive to instruction-based mitigation, likely reflecting tighter coupling between tool execution patterns and training-time conventions.

\subsection{Additional Results of Semantic Perturbations}
\label{app:sec:semantic}
We provide additional results on the interaction between semantic perturbations and execution settings in Appendix Figures~\ref{fig:by_noise_chinese_quad} and~\ref{fig:by_noise_hindi_quad}.
These figures complement the main-text analysis by illustrating how paraphrasing and synonym substitution affect error composition under different language–execution regimes.
Consistent with the trends discussed in Section~\ref{sec:exp:benchmarking-results}, semantic perturbations have a limited impact in fully translated settings, where execution failures are dominated by parameter value language mismatch.
In contrast, under partially translated queries or explicit English-parameter prompting, semantic perturbations introduce additional errors by disrupting the recovery of exact English parameter surface forms.
These results further support our conclusion that semantic noise primarily amplifies multilingual tool-calling failures when strict execution-level surface-form constraints are enforced.

\begin{figure*}[t]
  \centering
  \includegraphics[width=\textwidth]{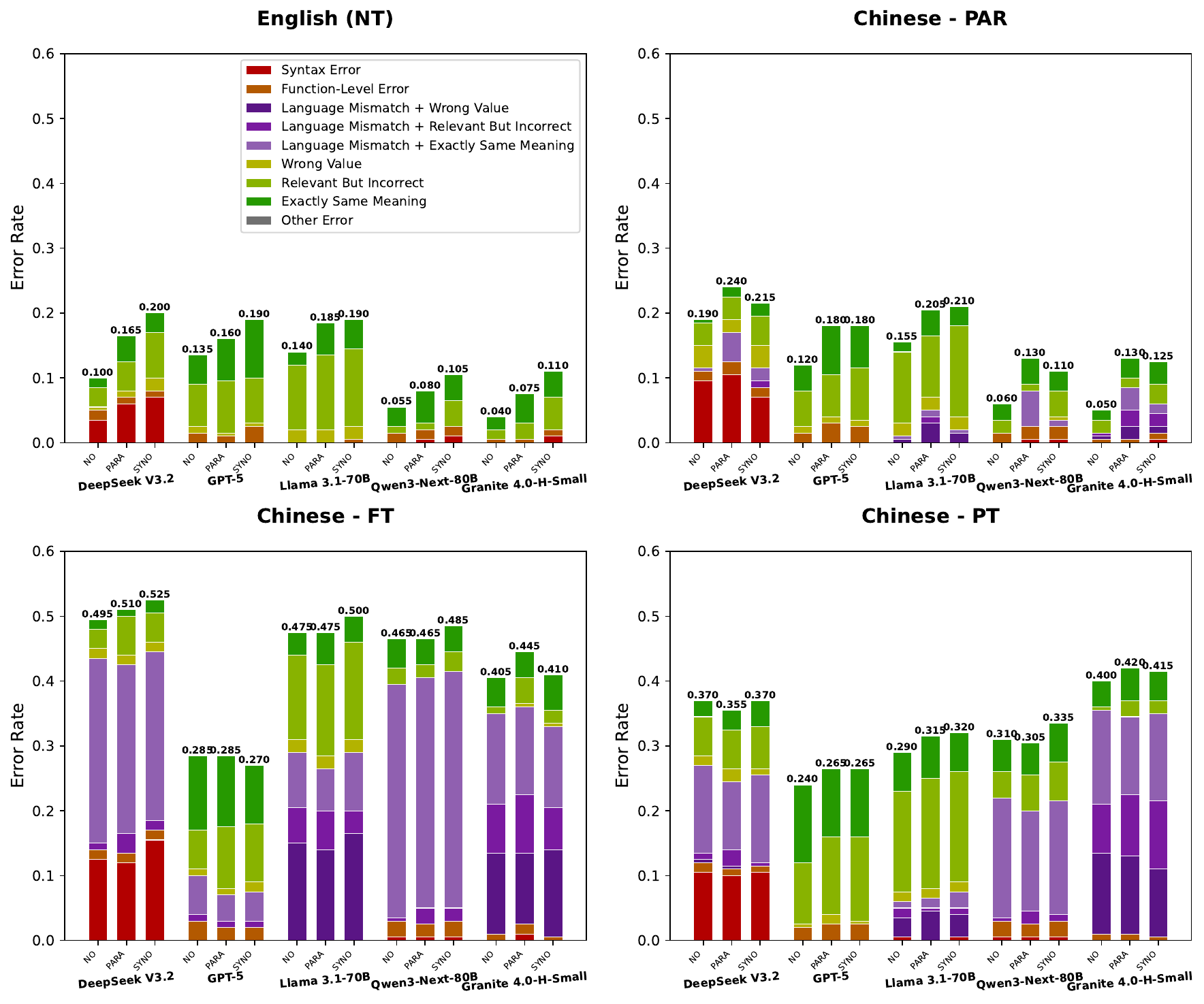}
  \caption{\textbf{Semantic perturbations primarily affect Chinese tool calling when English parameter surface forms are required.}
  We compare five representative models under semantic perturbations across three execution settings on Chinese queries: fully translated (FT), partially translated (PAR), and fully translated with explicit English-parameter prompting (PT).}
  \label{fig:by_noise_chinese_quad}
\end{figure*}

\begin{figure*}[t]
  \centering
  \includegraphics[width=\textwidth]{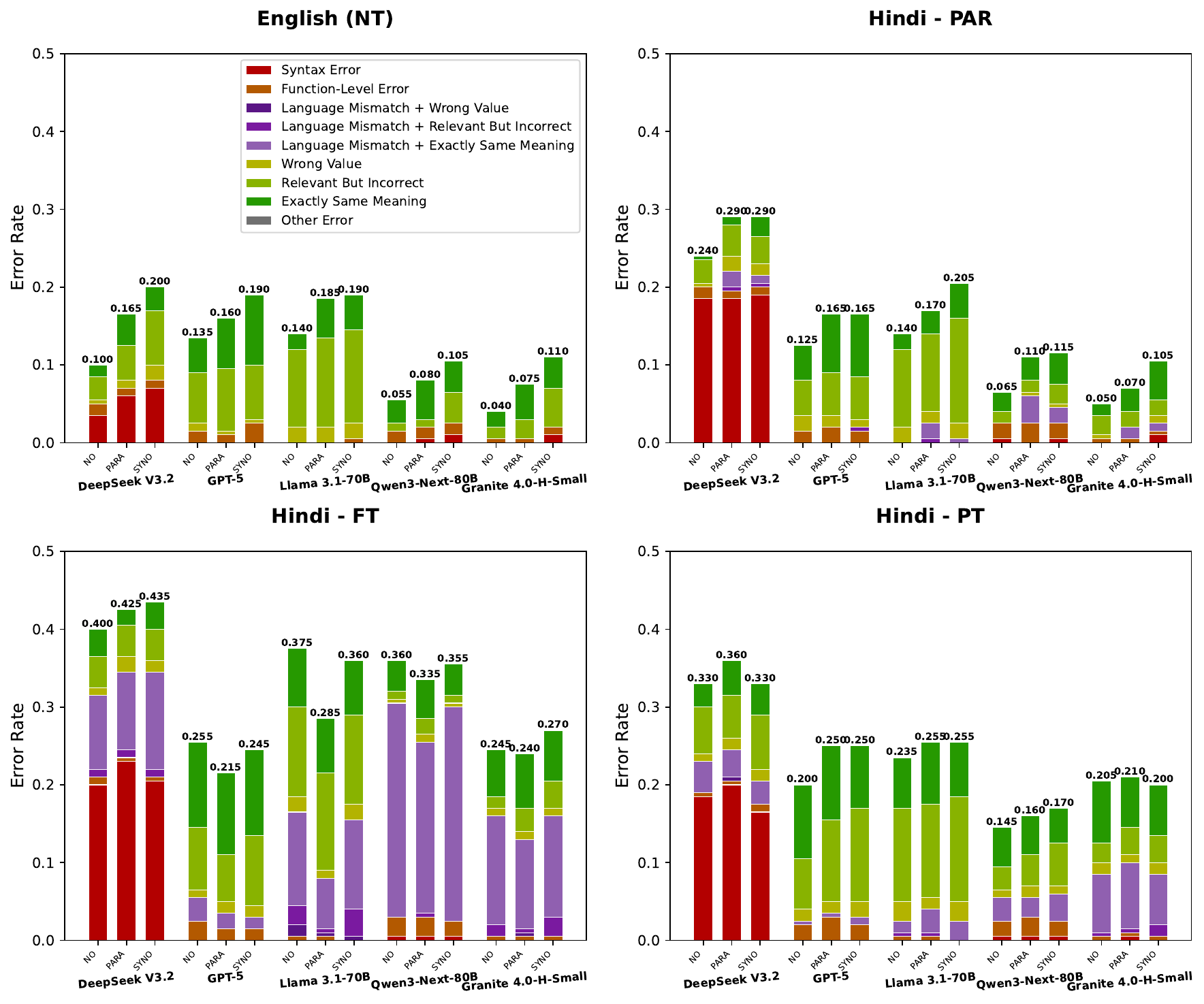}
\caption{\textbf{Semantic perturbations primarily affect Hindi tool calling when English parameter surface forms are required.}
  We compare five representative models under semantic perturbations across three execution settings: fully translated (FT), partially translated (PAR), and fully translated with English-parameter prompting (PT).}
  \label{fig:by_noise_hindi_quad}
\end{figure*}

\end{document}